\documentclass[10pt,twocolumn,letterpaper]{article}

\usepackage{iccv}
\usepackage{times}
\usepackage{epsfig}
\usepackage{graphicx}
\usepackage{amsmath}
\usepackage{amssymb}

\usepackage{subcaption}
\usepackage[table,xcdraw]{xcolor}
\usepackage{multirow}
\usepackage{adjustbox}

\usepackage[pagebackref=true,breaklinks=true,letterpaper=true,colorlinks,bookmarks=false]{hyperref}

\iccvfinalcopy 

\def\iccvPaperID{58} 

\ificcvfinal\fi

\begin{document}

\title{Deterministic Neural Illumination Mapping for Efficient Auto-White Balance Correction}

\author{Furkan Kınlı\textsuperscript{1} \qquad Doğa Yılmaz\textsuperscript{2} \qquad Barış Özcan\textsuperscript{3} \qquad Furkan Kıraç\textsuperscript{4} \\
Vision and Graphics Lab, Özyeğin University, Türkiye\\
{\tt\small \{furkan.kinli\textsuperscript{1}, furkan.kirac\textsuperscript{4}\}@ozyegin.edu.tr,} \\ {\tt\small \{doga.yilmaz.11481\textsuperscript{2}, baris.ozcan.10097\textsuperscript{3}\}@ozu.edu.tr}
}

\twocolumn[{%
\renewcommand\twocolumn[1][]{#1}%
\maketitle
\begin{center}
    \centering
    \captionsetup{type=figure}
    \includegraphics[width=0.86\textwidth]{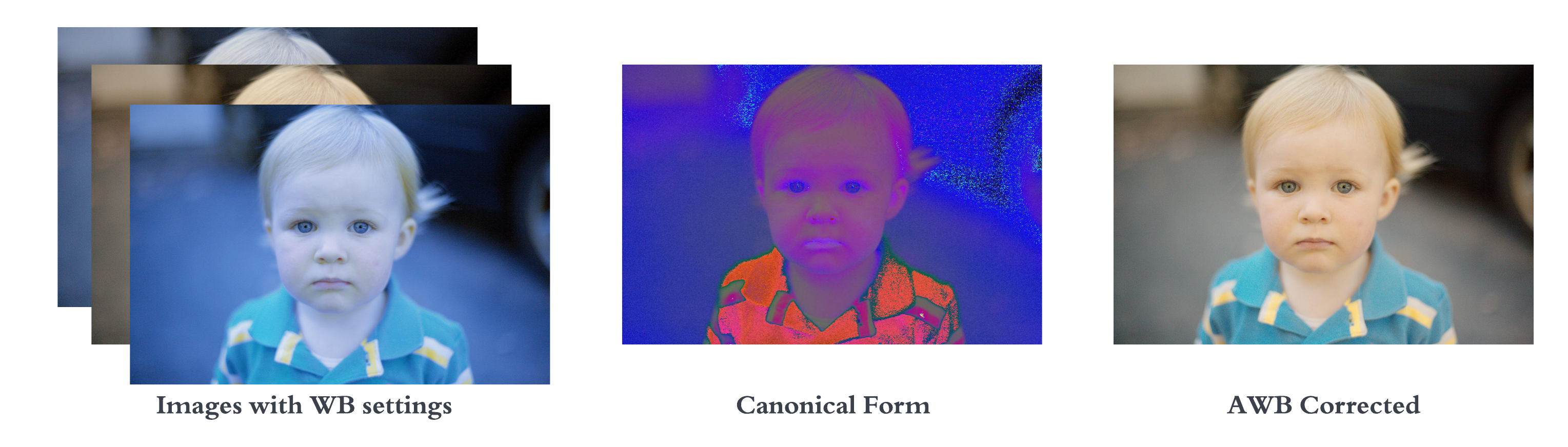}
    \caption{Learning deterministic illumination color mappings from images with different WB settings for both canonical illumination form and AWB corrected version. Pixel intensities in the canonical form are amplified for better visualization.}
    \label{fig:intro}
\end{center}%
}]

\ificcvfinal\thispagestyle{empty}\fi

\begin{abstract}
   Auto-white balance (AWB) correction is a critical operation in image signal processors for accurate and consistent color correction across various illumination scenarios. This paper presents a novel and efficient AWB correction method that achieves at least 35 times faster processing with equivalent or superior performance on high-resolution images for the current state-of-the-art methods. Inspired by deterministic color style transfer, our approach introduces deterministic illumination color mapping, leveraging learnable projection matrices for both canonical illumination form and AWB-corrected output. It involves feeding high-resolution images and corresponding latent representations into a mapping module to derive a canonical form, followed by another mapping module that maps the pixel values to those for the corrected version. This strategy is designed as resolution-agnostic and also enables seamless integration of any pre-trained AWB network as the backbone. Experimental results confirm the effectiveness of our approach, revealing significant performance improvements and reduced time complexity compared to state-of-the-art methods. Our method provides an efficient deep learning-based AWB correction solution, promising real-time, high-quality color correction for digital imaging applications. Source code is available at \url{https://github.com/birdortyedi/DeNIM/}
\end{abstract}

\section{Introduction}

In the realm of digital imaging, auto-white balance (AWB) correction is one of the most critical operations in image signal processors (ISPs). The colors presented in the final sRGB image should be somehow aligned with the colors perceived by the human eye. This operation mainly aims to ensure accurate and consistent color correction across a variety of illumination scenarios. Due to the effect of differing light sources in real-world scenarios, which possess continuous range of color temperatures, AWB correction task still remains challenging. Recent studies on AWB correction generally introduce a method to model leading illumination settings and undesired color casts in the scene, and then subsequently adjust the color balance. 

\begin{figure*}[!ht]
  \centering
  \includegraphics[width=\textwidth]{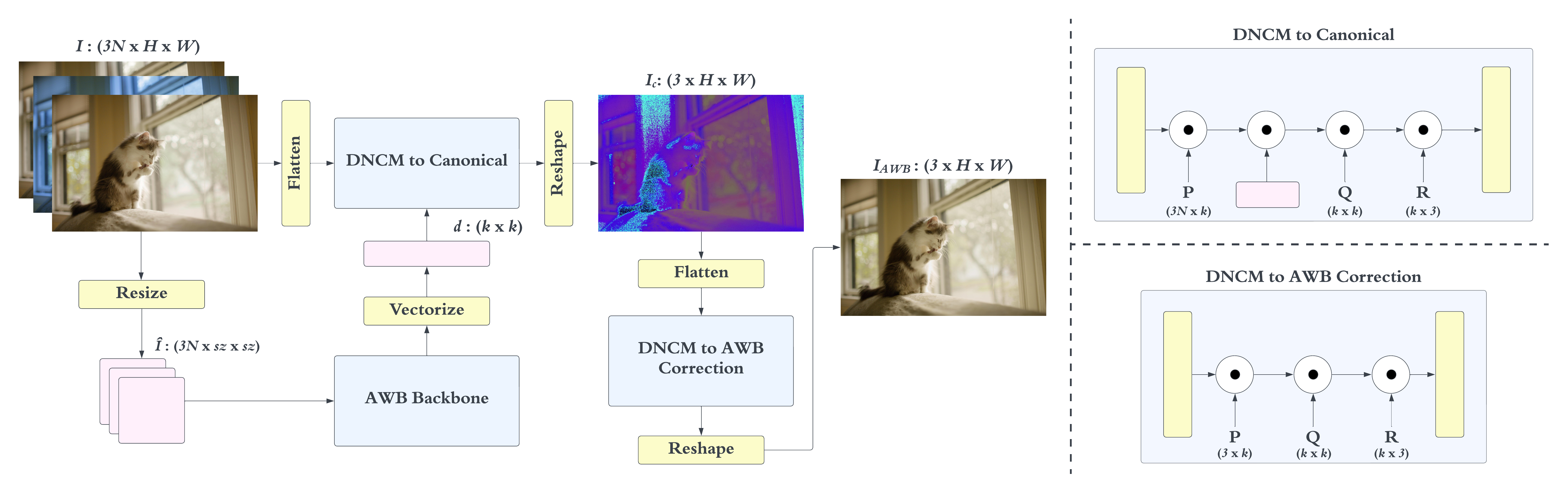}
  \caption{Overall design of our proposed illumination mapping strategy. We first reduce the resolution of the input images to a compatible size with the AWB correction backbone (\textit{i.e.}, Mixed WB \cite{Afifi_2022_WACV}, Style WB \cite{Kinli_2023_WACV}). Then, high-resolution input images and the latent representations of low-resolution versions, extracted by AWB correction backbone, are fed into a deterministic color mapping module (DNCM) \cite{Ke_2023_CVPR} to obtain a canonical form. Another DNCM module (without fusion capability) takes the canonical form as input and learns to map the pixel values to the ones for AWB corrected version. This strategy ensures that the AWB correction model is resolution-agnostic.}
  \label{fig:arc_figure}
\end{figure*}

A number of AWB correction methods have been introduced, which employ various strategies (\textit{e.g.}, low-level statistical methods, gamut-based methods, and learning-based methods). Earlier studies \cite{buchsbaum1980spatial,brainard1986analysis,finlayson2004shades,van2007edge,gijsenij2011improving,cheng2014illuminant,joze2012role,qian2018revisiting,qian2019finding} benefit from low-level statistics of images or patches to infer the illumination, and employ a simple diagonal-based correction matrix \cite{gijsenij2011computational} of predicted illumination to rectify the color casts in the scene. In addition to low-level statistical methods, gamut-based methods \cite{forsyth1990novel,finlayson2000improving,finlayson2006gamut,gijsenij2010generalized} mainly introduce models that aims to learn mappings from the images captured under unknown lighting conditions to the reference colors captured under known lighting conditions. Learning-based methods \cite{brainard1994bayesian,funt1996learning,brainard1997bayesian,gehler2008bayesian,hernandez2020multi} have become more popular when compared to their ancestors, due to their better capability of representing the illumination in real-world scenarios.


With the advancements in computational photography, deep learning-based methods \cite{lou2015color,shi2016deep,Bianco_2019_CVPR,Hu_2017_CVPR,xu2020end,lo2021clcc,afifi2021cross,Afifi_2022_WACV,ono2022degree,Kinli_2023_WACV,li2023mimt} have demonstrated an outstanding performance edge over all previous AWB correction strategies. However, the high computational requirements and significant power demands of these approaches restrict their direct integration within a camera pipeline. Especially, the recent approaches suffer from the computational complexity mostly leading to better performance without considering the time efficiency and their practical usage. Addressing this issue, we propose a novel, deep learning-based AWB correction method, which makes the current state-of-the-art methods at least 35 times faster, while delivering equivalent or better performance on high-resolution images.

The main contributions of this study can be summarized as follows:

\begin{itemize}
    \item We propose a novel and efficient strategy for AWB correction, which learns deterministic color mappings for both canonical illumination and AWB-corrected forms with the help of learnable projection matrices.
    \item Our design allows the input to be resolution-agnostic and any pre-trained AWB network can be integrated into this design as the backbone network.
    \item We demonstrate that employing deterministic illumination color mapping for AWB correction yields a substantial improvement in the performance of existing state-of-the-art methods, while significantly reducing the time complexity, achieving a speedup of at least 35 times faster.
\end{itemize}


\section{Methodology}

Given a set of high-resolution images with different white balance (WB) settings $I$, our proposed strategy learns to achieve a deterministic illumination color mapping for efficient AWB correction. Prior works \cite{Afifi_2022_WACV, Kinli_2023_WACV} focus on learning the weighting maps for all different WB settings in low-resolution space. Then, they render the AWB-corrected version in high-resolution space by linearly combining images with different WB settings and their corresponding weighting maps. Although this approach can produce quite well outputs, it essentially requires multi-scale inference and smoothing after resizing the weighting maps to the original resolution to significantly improve the results. However, these post-processing steps make this approach challenging to use in practical scenarios.

Inspired by deterministic color style transfer \cite{Ke_2023_CVPR}, we developed an idea of deterministic illumination color mapping for AWB correction. The overall design of our proposed illumination mapping strategy is shown in Figure \ref{fig:arc_figure}. First, we reduce the resolution of the input images $I$ to make them compatible with the architectures of prior works (\textit{i.e.}, $256$px). By using only the encoder part of one of these architectures, we feed low-resolution images $\hat{I}$ into the encoder to extract rich information from different WB settings. To obtain the latent representations, we use $1 \times 1$ convolutional layer followed by GeLU activation \cite{hendrycks2016gaussian} to vectorize the feature maps. This provides an image-adaptive color mapping matrix $d$ for DNCM module \cite{Ke_2023_CVPR} to generate a canonical form. 

\begin{equation}
    \mathit{d}^{(k \times k)} = V(E(\mathbf{\hat{I}}))
\end{equation}
where $E$ refers to the AWB encoder (\textit{i.e.}, \cite{Afifi_2022_WACV} or \cite{Kinli_2023_WACV}), $V$ stands for the vectorization operation by $1\times1$ convolutional layer and activation. Note that we use pre-trained weights for $E$ and freeze its weights during our training.

For \textit{DNCM to canonical} module, the first step involves unfolding high-resolution image $I$ into a 2D matrix of dimensions ($HW \times 3N$), where $N$ refers to the number of WB settings, $H$ and $W$ represent height and width, respectively. Each pixel in $I$ is then transformed into a $k$-dimensional vector using a projection matrix $P$ ($3N \times k$). $k$ can be any number depending on the computational power, but we set it to $32$ in our design. The extracted image-adaptive color mapping matrix $d$ is multiplied with $k$-dimensional vector to inject the rich information into the projected space. $Q$ ($k \times k$) and $R$ ($k \times 3$) are the following learnable projection matrices to form the canonical form in this module. We can formulate this module, namely \textit{DNCMc}, as follows

\begin{equation}
    DNCMc(\mathbf{I}, \mathit{d}) = \mathbf{I}^{(HW \times 3)} \cdot \mathbf{P}^{(3 \times k)} \cdot \mathit{d}^{(k \times k)} \cdot \mathbf{Q}^{(k \times k)} \cdot \mathbf{R}^{(k \times 3)}
\end{equation}
where $\cdot$ denotes the matrix multiplication.

Next, we feed the canonical form into \textit{DNCM to AWB correction} module (\textit{DNCMa}). It does not have any fusion capability but learns to directly map the pixel values in the canonical form to the correct ones for the AWB version. Each pixel in the canonical form $I_c$ is transformed into a $k$-dimensional vector by a projection matrix $P$ ($3 \times k$). By using a similar design to \textit{DNCMc}, $Q$ ($k \times k$) and $R$ ($k \times 3$) are responsible for converting the embedded $k$-dimensional vector back to the RGB color space, which finally forms the output $I_{AWB}$. The formal definition of \textit{DNCMa} can be seen in Equation \ref{eq:dncma}.

\begin{equation}
    \label{eq:dncma}
    DNCMa(\mathbf{I_c}) = \mathbf{I_c}^{(HW \times 3)} \cdot \mathbf{P}^{(3 \times k)}  \cdot \mathbf{Q}^{(k \times k)} \cdot \mathbf{R}^{(k \times 3)}
\end{equation}

Apart from the self-supervised learning mechanism for DNCM, followed in \cite{Ke_2023_CVPR}, the learning objective is to minimize the reconstruction error between the ground truth and the AWB corrected output, as shown in Equation \ref{eq:loss}.
\begin{equation} 
    \label{eq:loss}
    \mathcal{L} = || \mathbf{I_{GT}} - \mathbf{I_{AWB}} ||^2_F     
\end{equation}
where $I_{GT}$ and $I_{AWB}$ denote the ground truth image and the output. To keep the training process simple and tractable, we did not include the smoothing loss \cite{Afifi_2022_WACV} or perceptual loss \cite{johnson2016perceptual} in our final objective function.

Our design removes the decoder part that generates the weighting maps in the prior works, and instead, it directly computes the illumination color mapping with two distinct DNCM modules for the canonical form and AWB-corrected version. This design mitigates the need for further post-processing of the weighting maps, which leads to reducing the time complexity without compromising the performance. Moreover, due to the one-by-one pixel value mapping characteristic delivered by matrix multiplications, it gives AWB correction model the ability to be resolution-agnostic. Lastly, any AWB correction method can be easily plugged into this design for extracting rich information in low-resolution space from different WB settings, which makes our design also model-agnostic.

\section{Experiments}

\subsection{Experimental Details}

For our training, we have employed the RenderedWB dataset \cite{Afifi_2022_WACV}, which contains 65,000 sRGB images with pre-defined WB settings and corresponding white-balanced versions, captured by different cameras. Following the experimental setup in the prior works, we have two sets of pre-defined WB settings, which are \texttt{\{t,f,d,c,s\}} and \texttt{\{t,d,s\}}. The color temperatures used for pre-defined WB settings are as follows: Tungsten (\texttt{t}, 2850K), Fluorescent (\texttt{f}, 3800K), Daylight (\texttt{d}, 5500K), Cloudy (\texttt{c}, 6500K), and Shade (\texttt{s}, 7500K). We did not apply any data augmentation technique to the images during our training. 

For all experiments, we freeze the weights of the AWB backbone and only trained DNCM modules in our proposed strategy from scratch. We set the size of the low-resolution space to $256$. We used AdamW optimizer \cite{loshchilov2017decoupled} ($\beta_1 = 0.9$, $\beta_2 = 0.999$) with batch size of 16. The learning rate is set to $1e-4$ and we did not employ any scheduling strategy. We did not apply any post-processing operations after obtaining the output.

\subsection{Evaluation}

Following the prior works \cite{Afifi_2020_CVPR, Afifi_2022_WACV, Kinli_2023_WACV}, we evaluate the AWB correction quality in terms of the mean-squared error (MSE), mean angular error (MAE) and color difference ($\Delta$E 2000). We report the mean, first (Q1), second (Q2), and third (Q3) quantile averages for all metrics.

\begin{figure*}[ht!]
    \centering
    \begin{subfigure}{0.24\textwidth}
        \includegraphics[width=\textwidth]{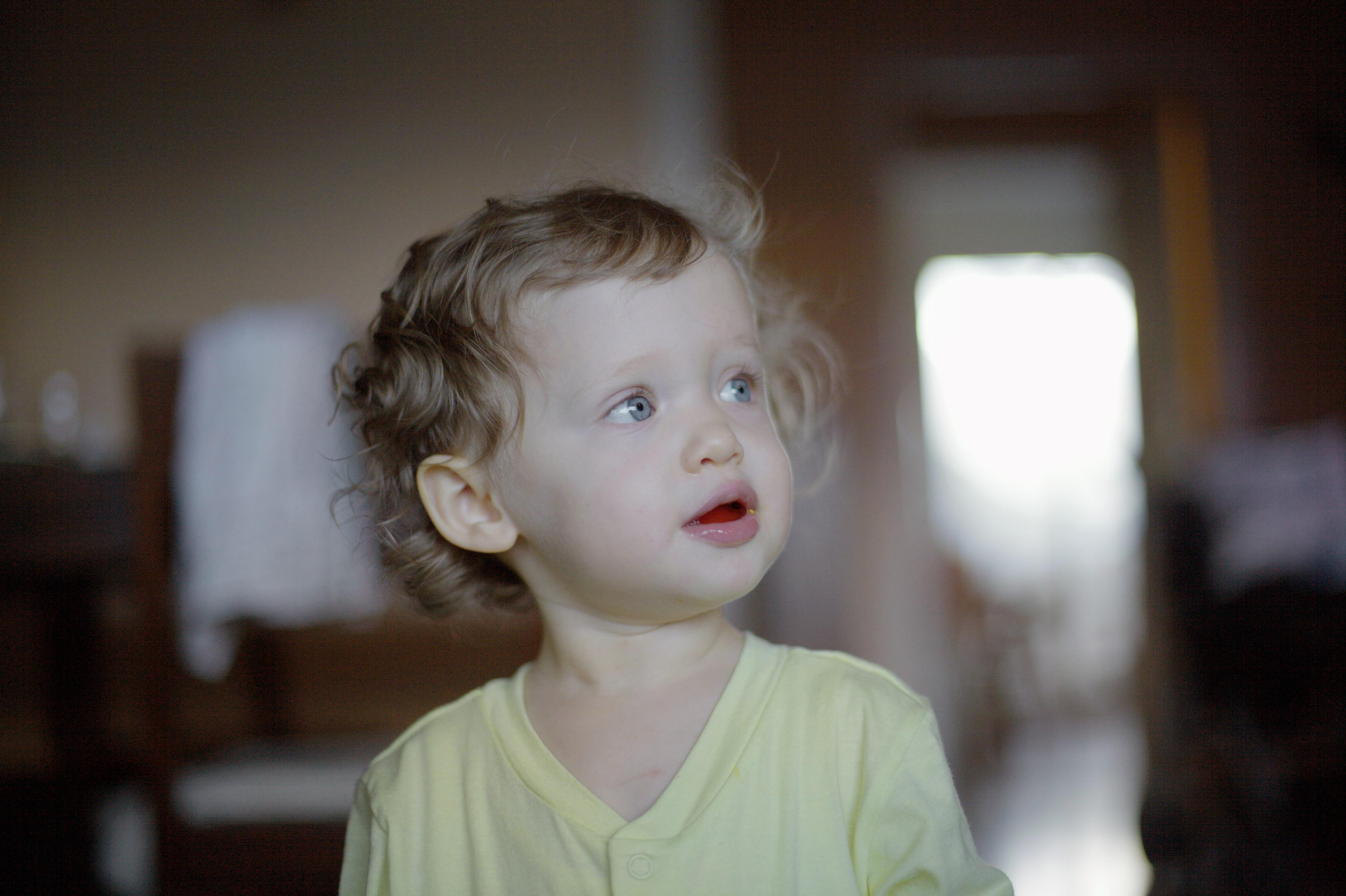}
        \includegraphics[width=\textwidth]{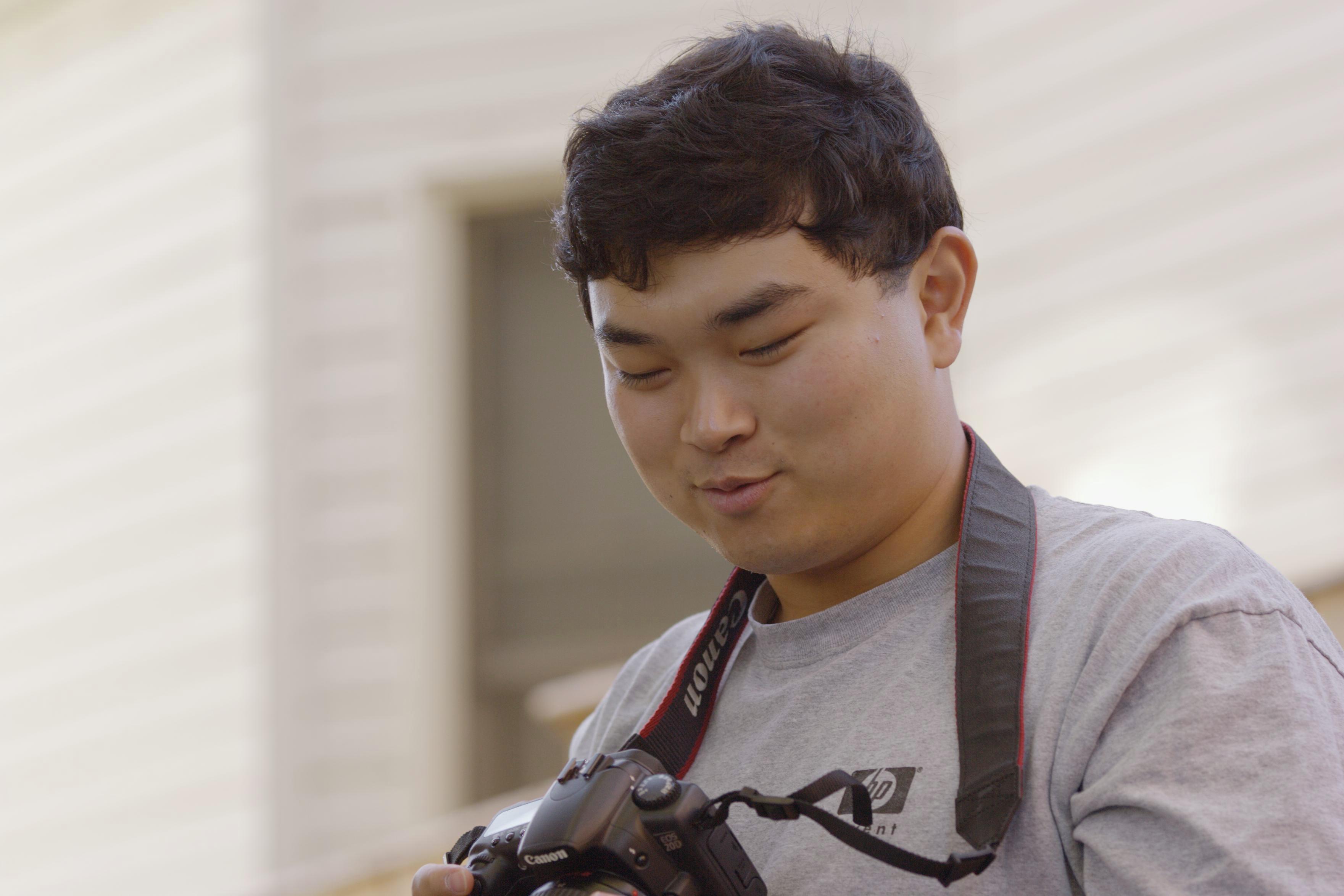}
        \includegraphics[width=\textwidth]{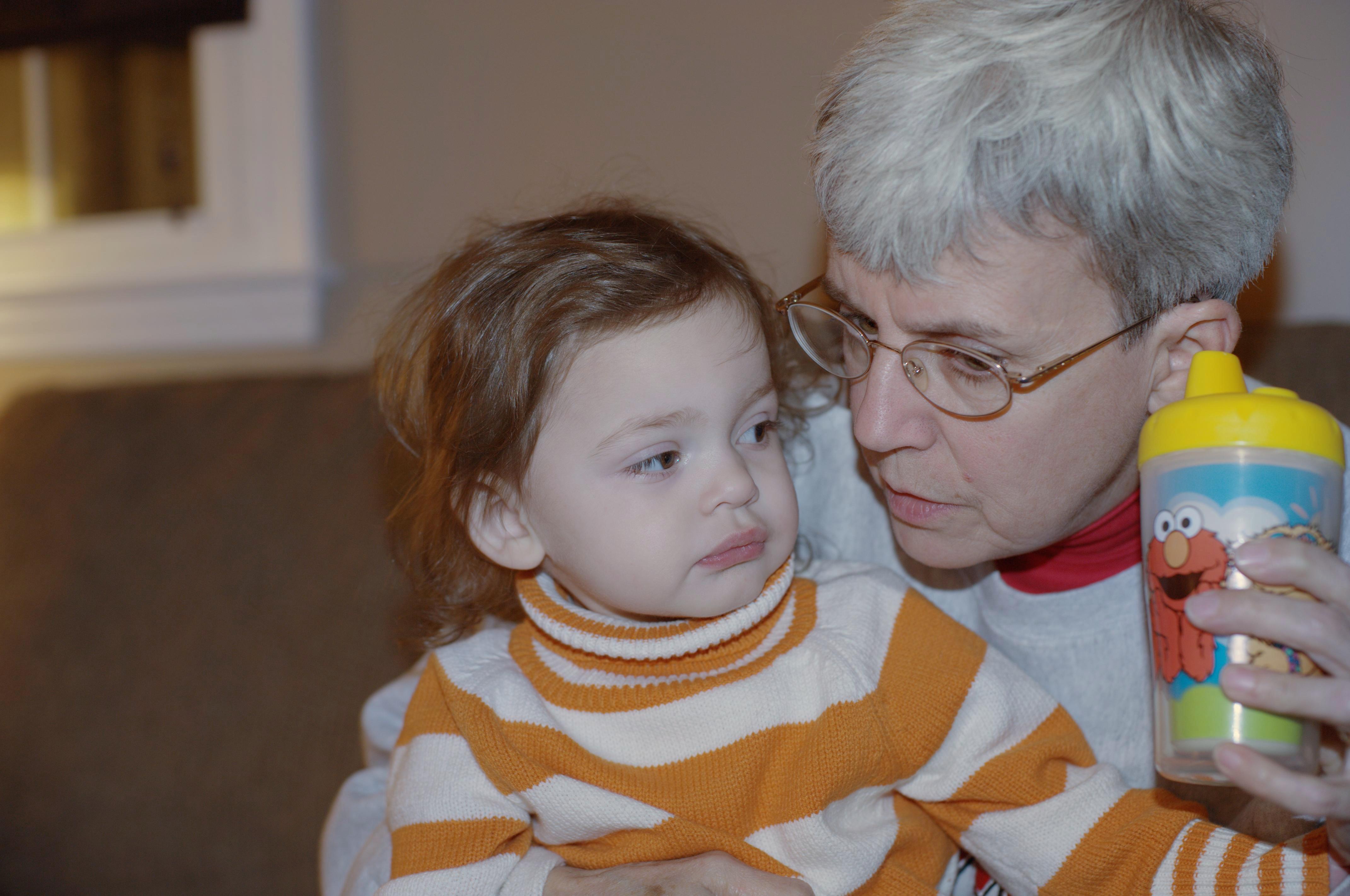}
        \includegraphics[width=\textwidth]{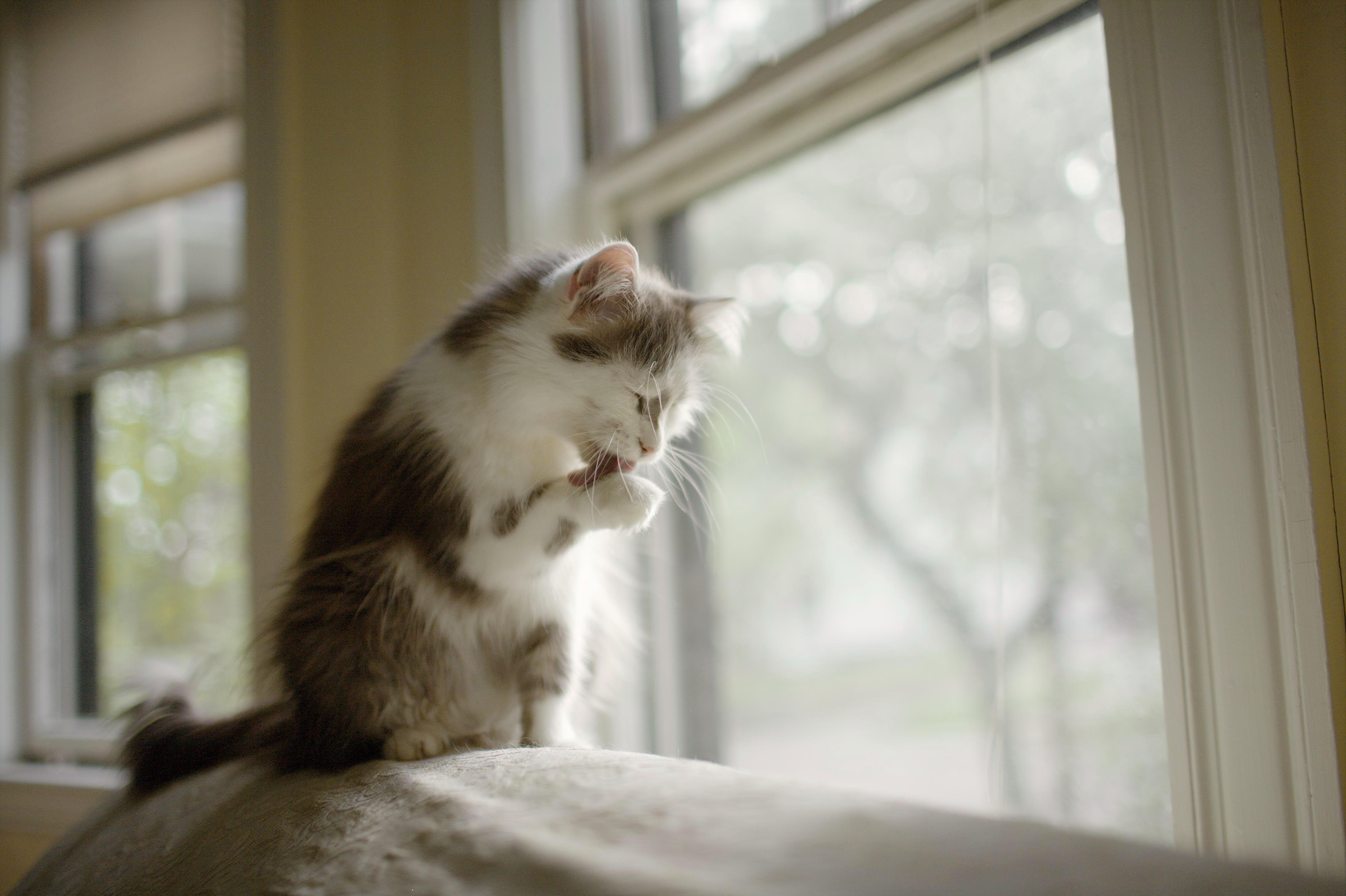}
        \includegraphics[width=\textwidth]{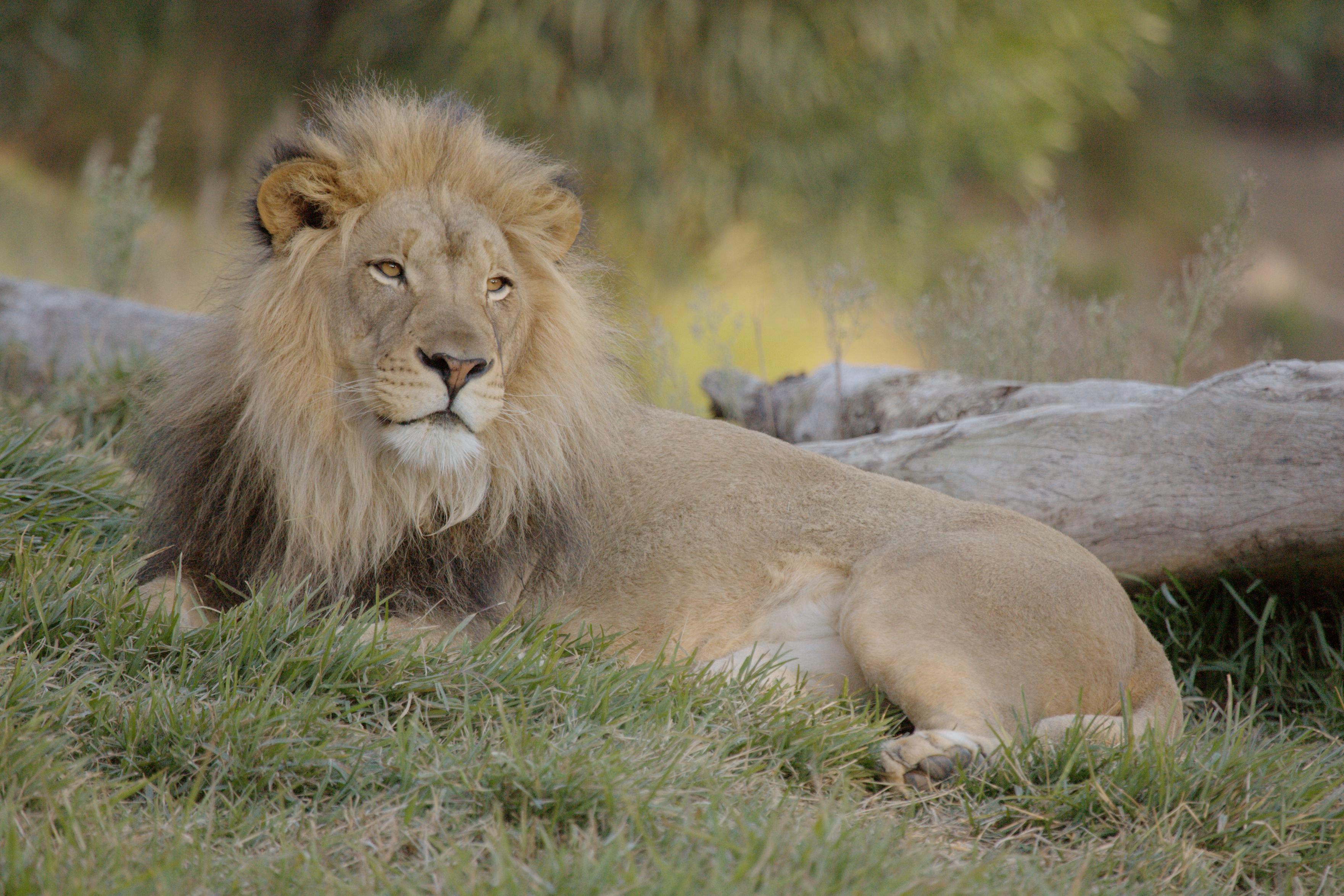} 
        \caption{Mixed WB}
    \end{subfigure}
    \begin{subfigure}{0.24\textwidth}
        \includegraphics[width=\textwidth]{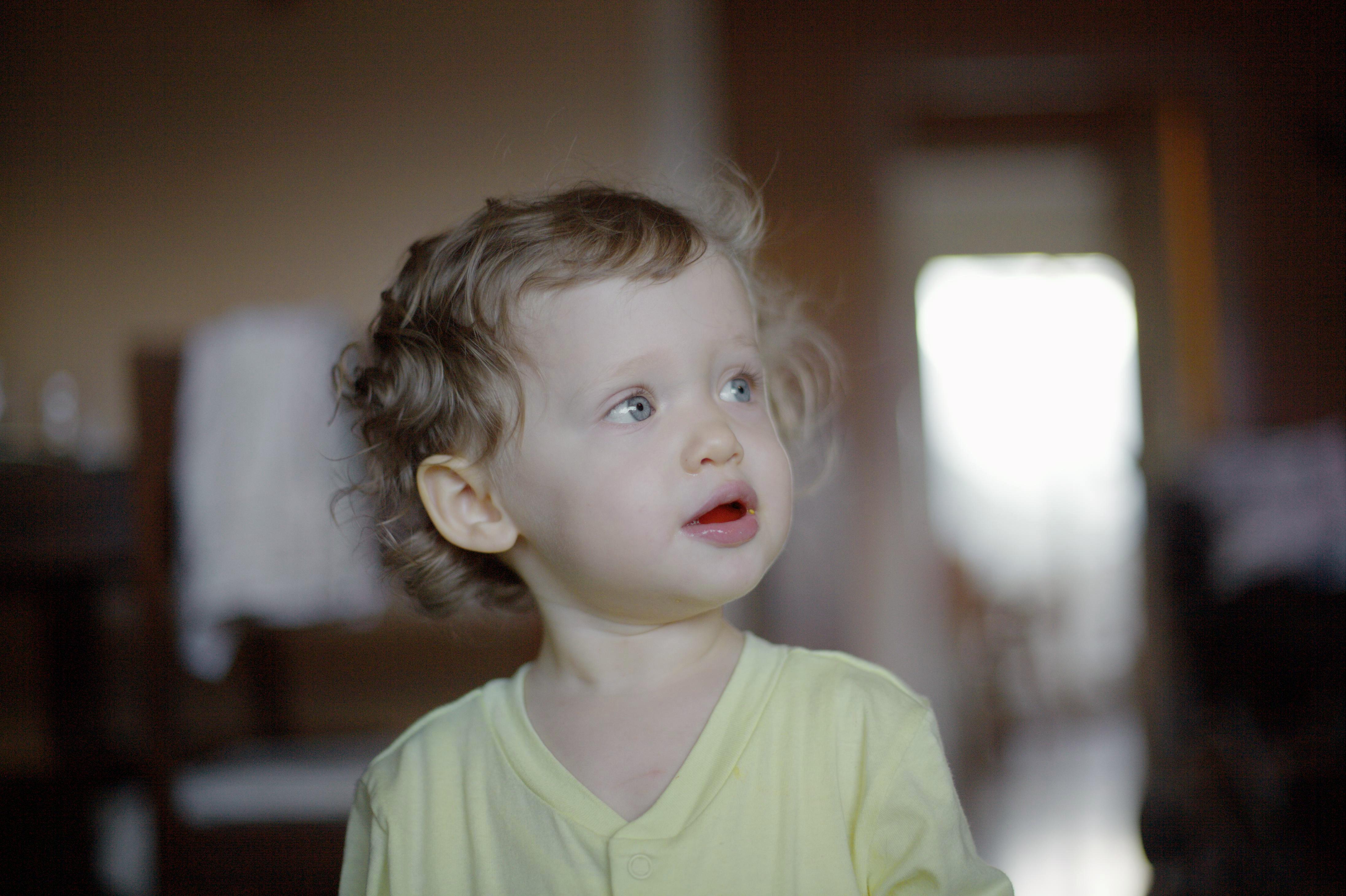}
        \includegraphics[width=\textwidth]{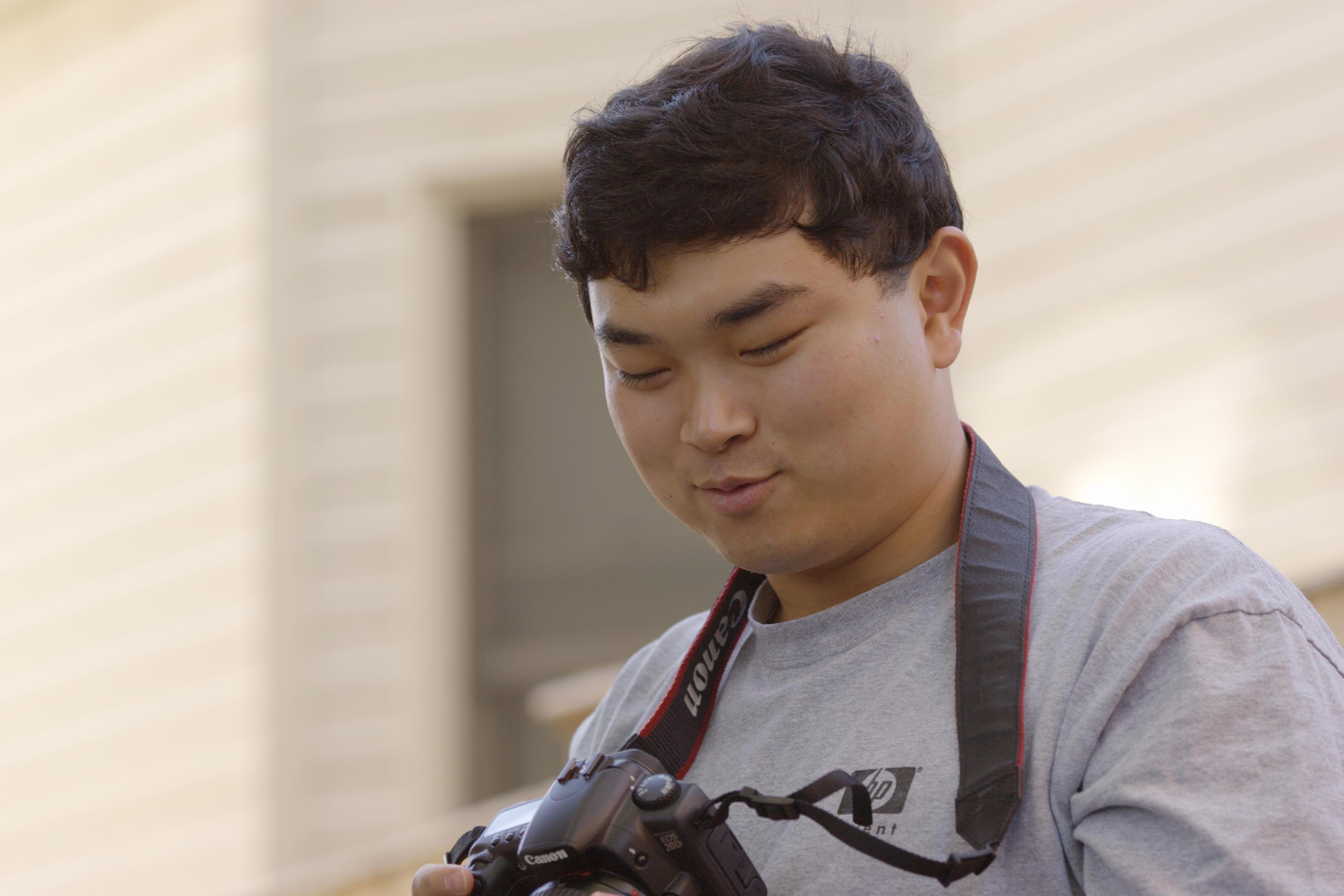}
        \includegraphics[width=\textwidth]{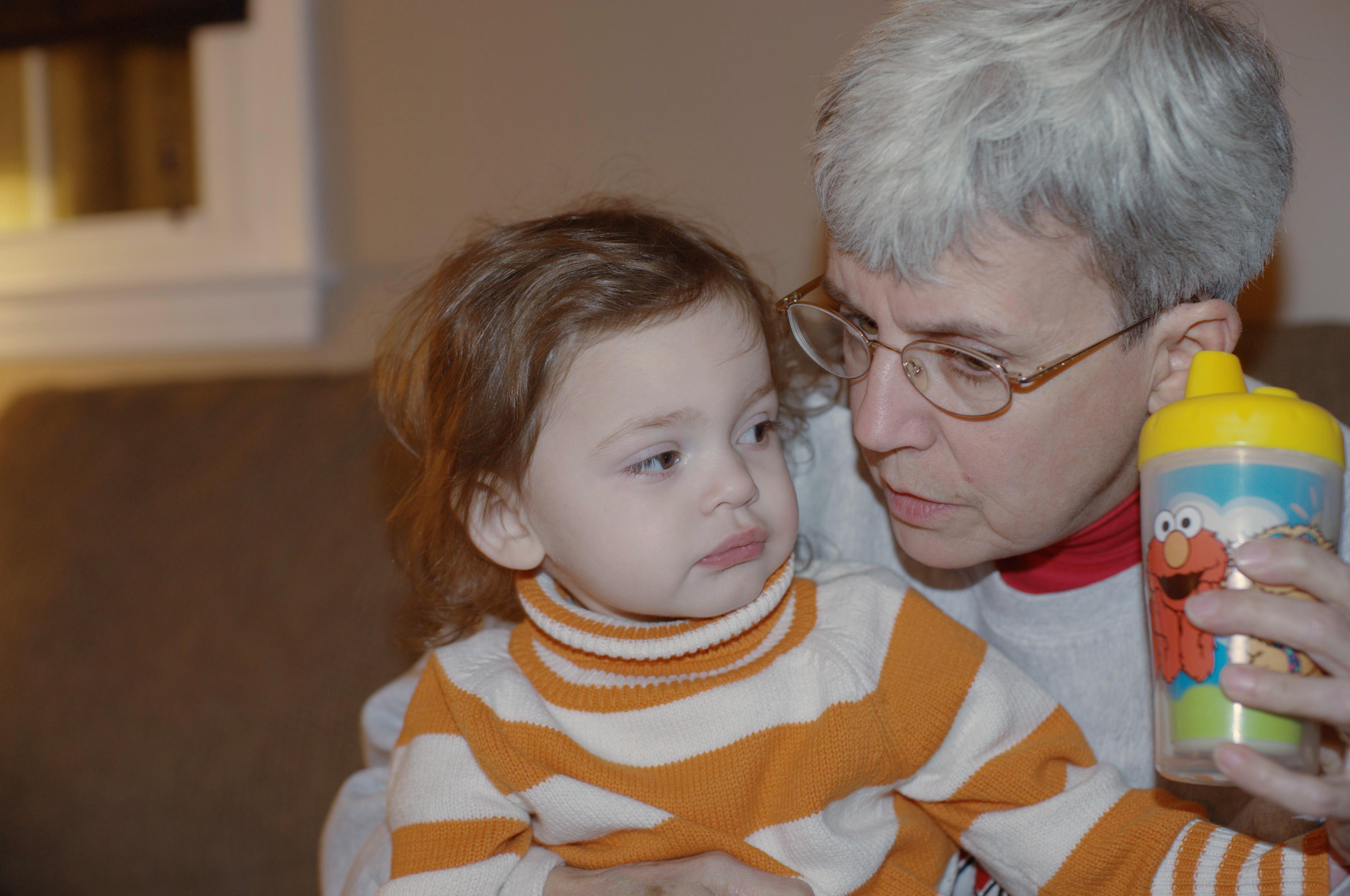}
        \includegraphics[width=\textwidth]{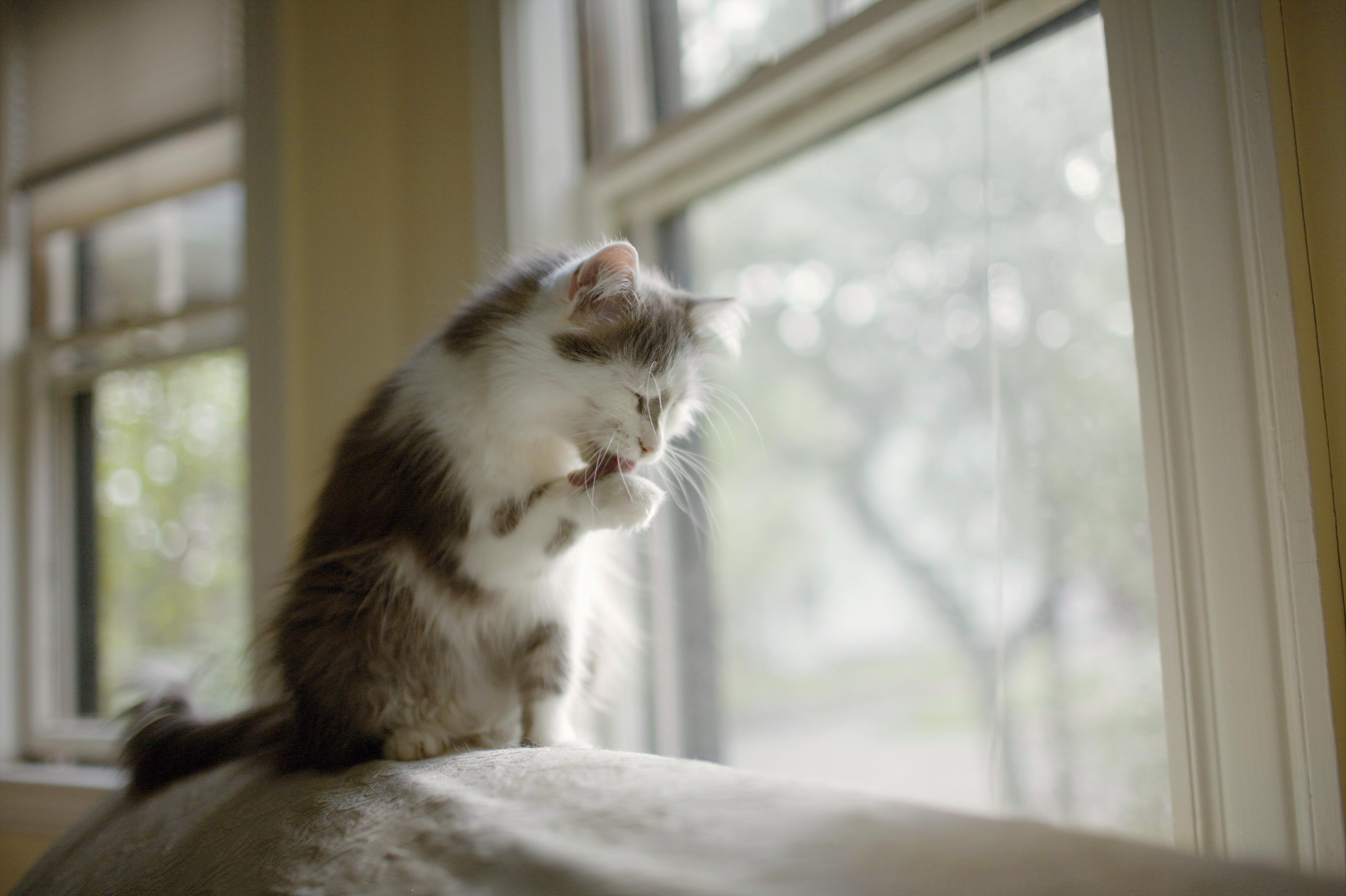}
        \includegraphics[width=\textwidth]{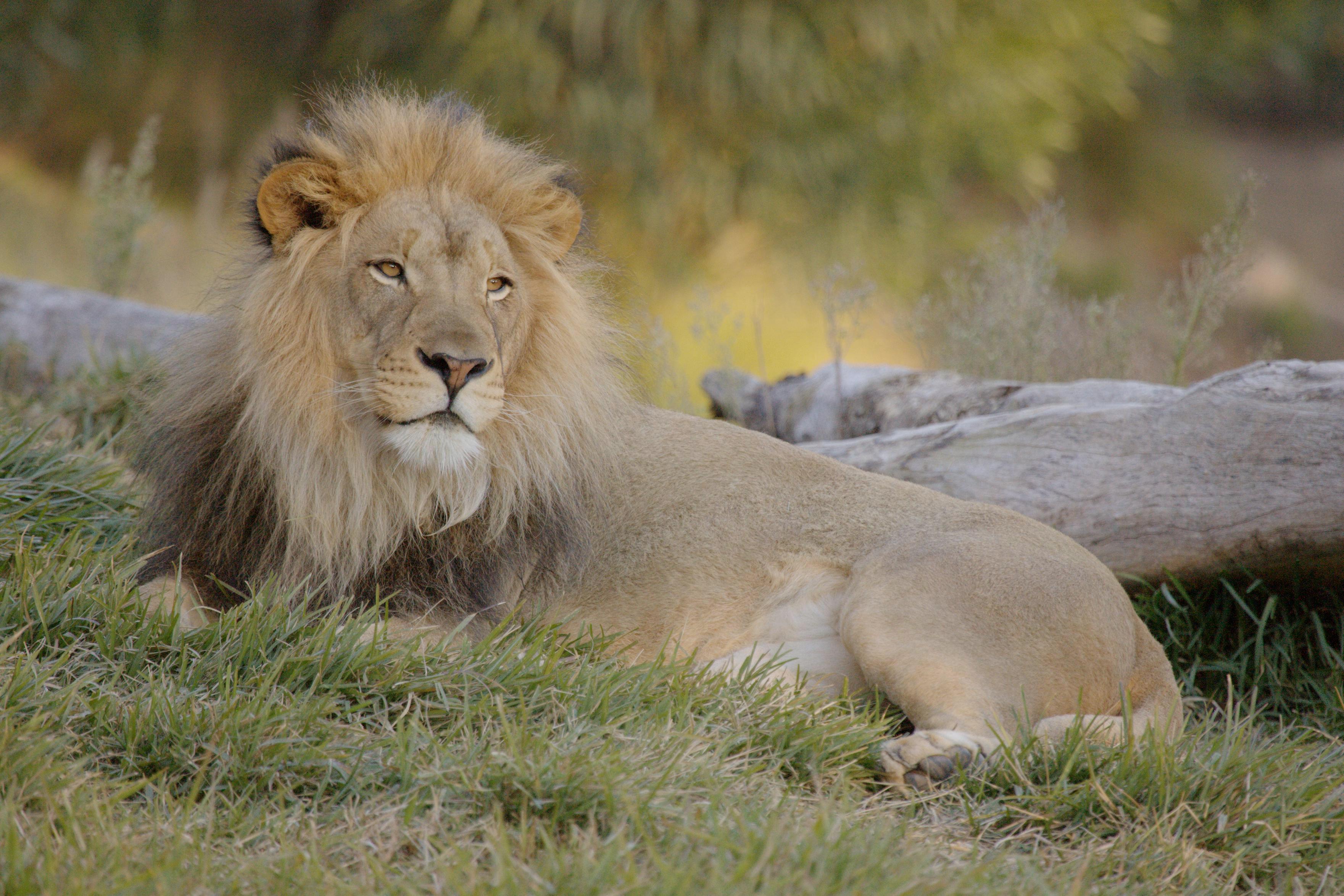} 
        \caption{Style WB}
    \end{subfigure}
    \begin{subfigure}{0.24\textwidth}
        \includegraphics[width=\textwidth]{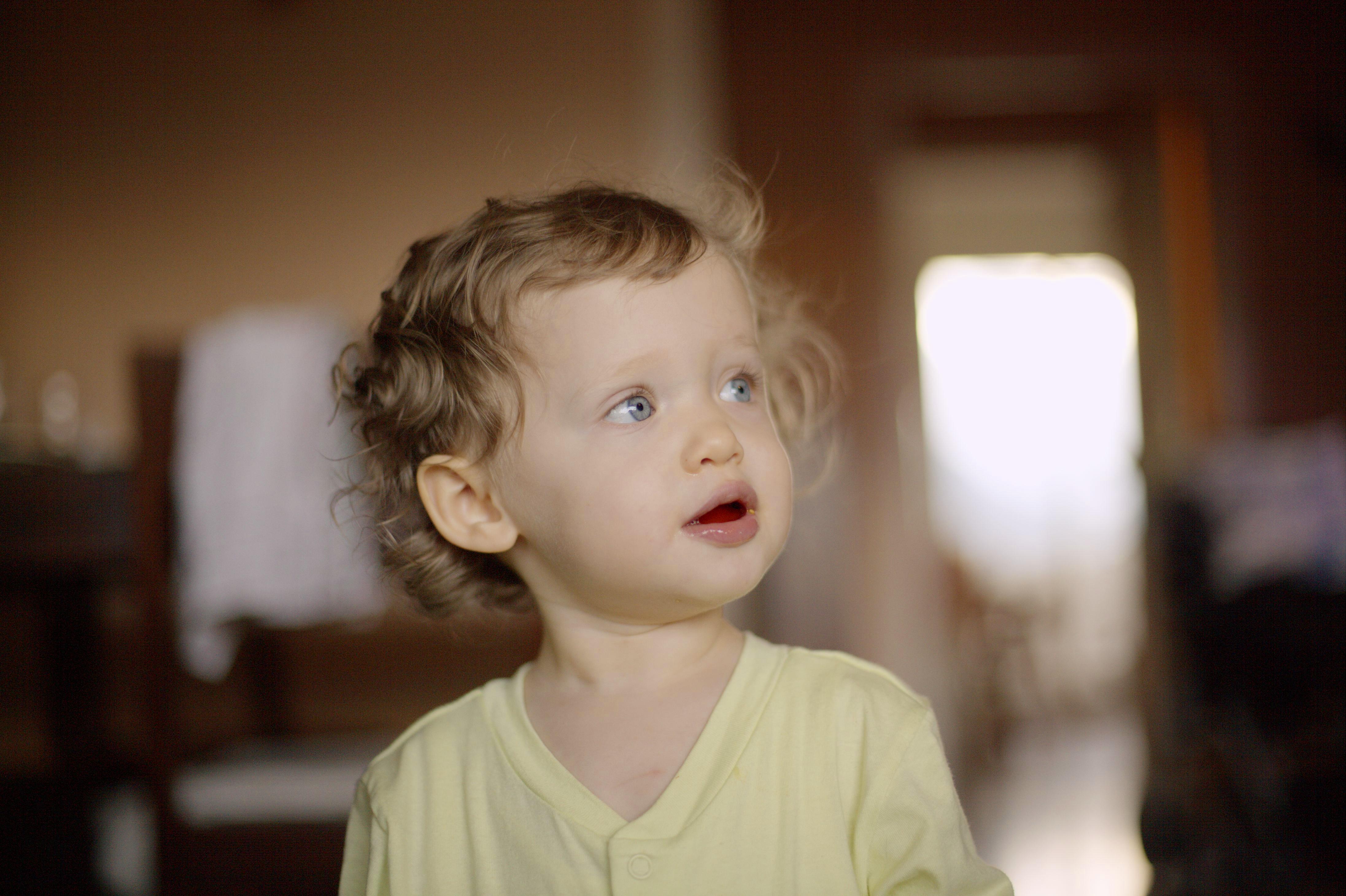}
        \includegraphics[width=\textwidth]{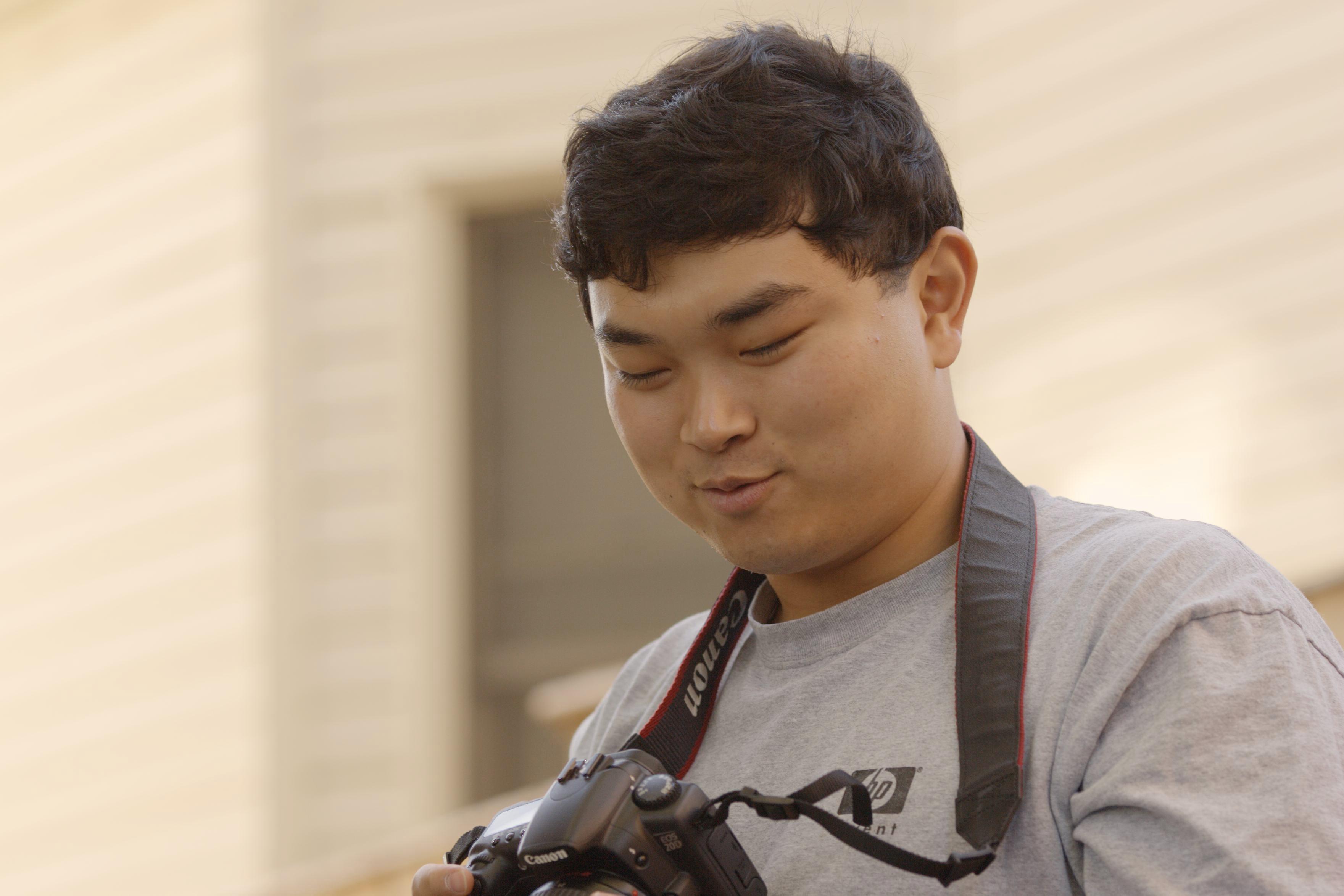}
        \includegraphics[width=\textwidth]{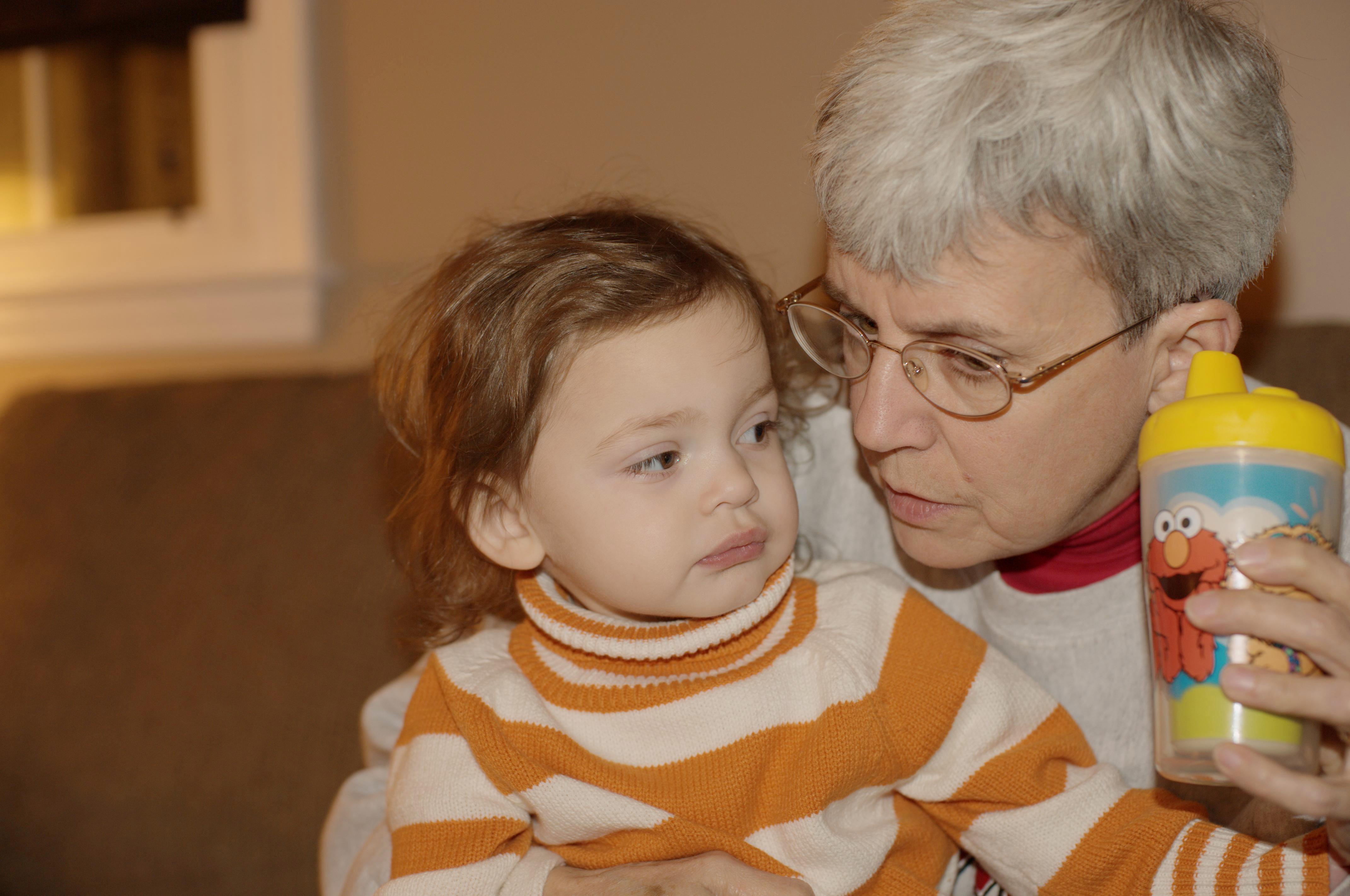} 
        \includegraphics[width=\textwidth]{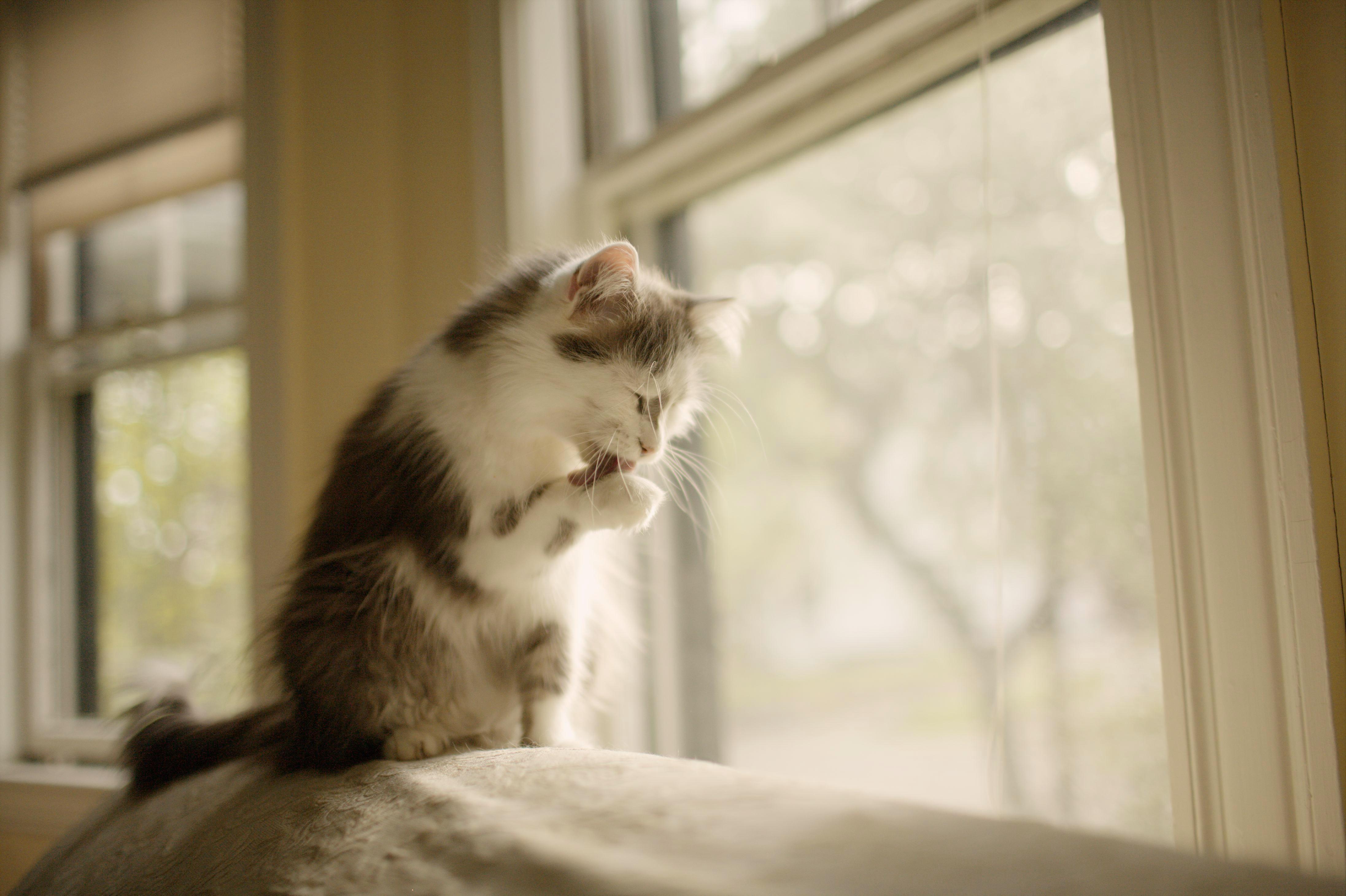} 
        \includegraphics[width=\textwidth]{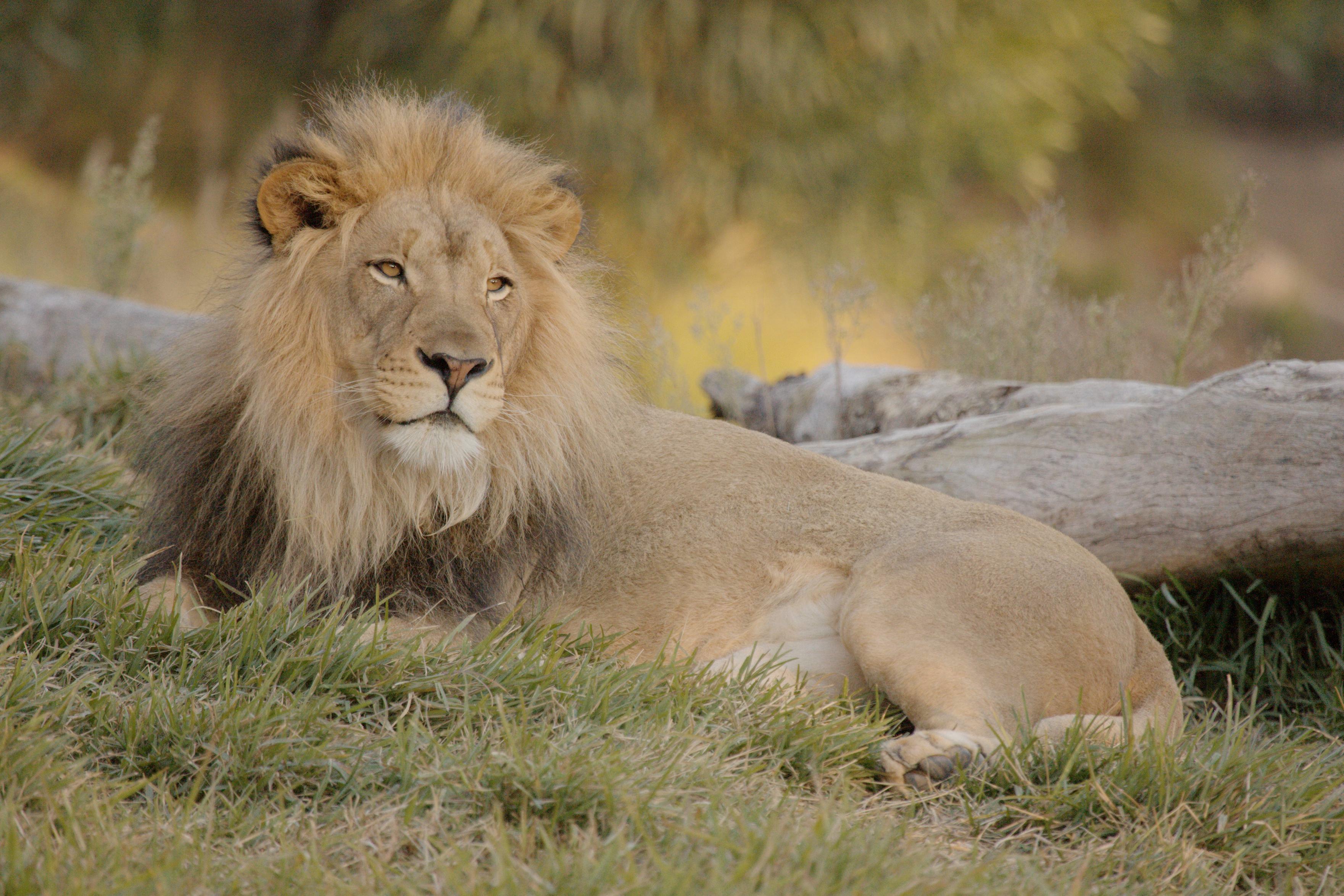} 
        \caption{DeNIM + Mixed WB}
    \end{subfigure}
    \begin{subfigure}{0.24\textwidth}
        \includegraphics[width=\textwidth]{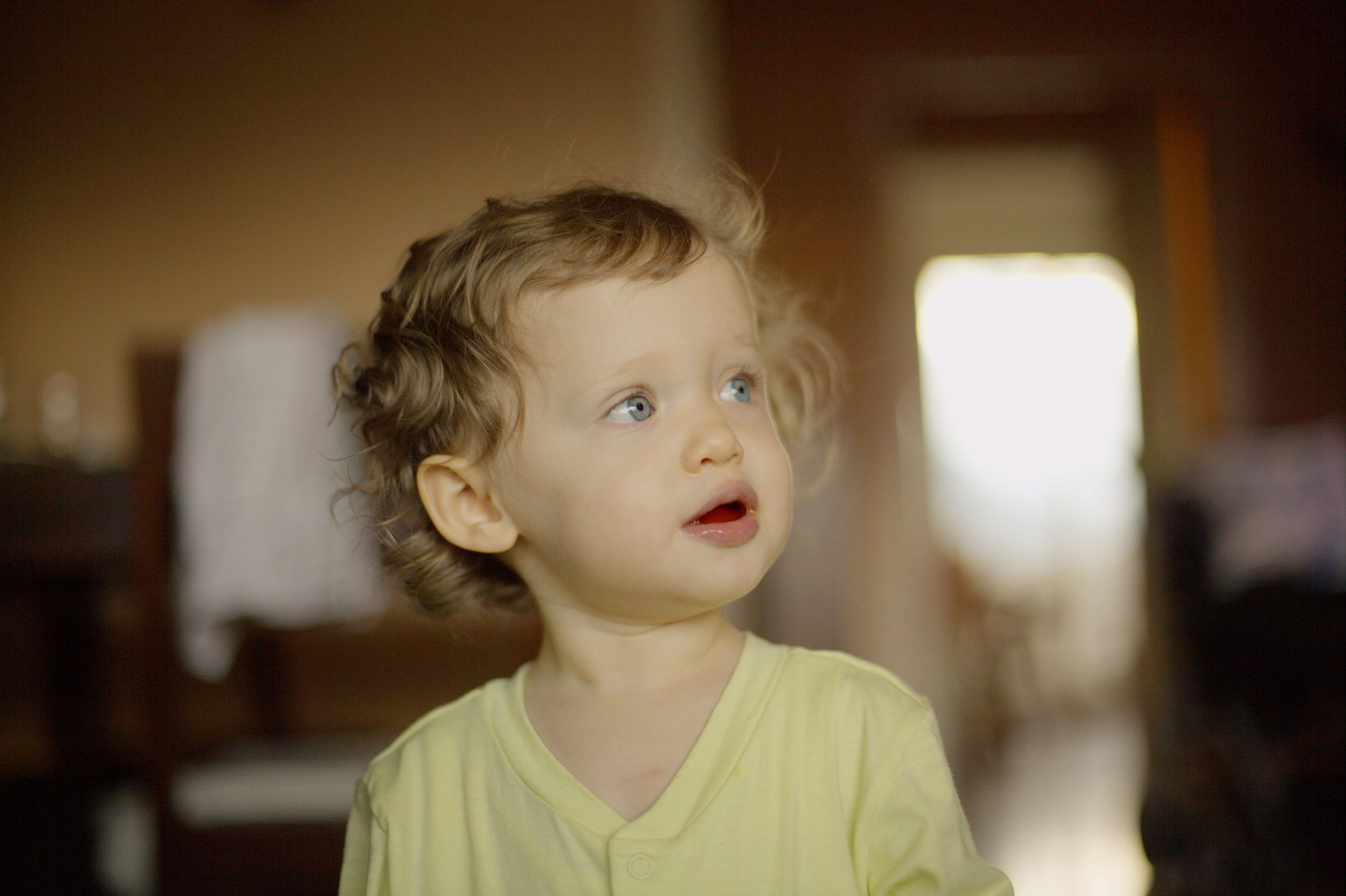}
        \includegraphics[width=\textwidth]{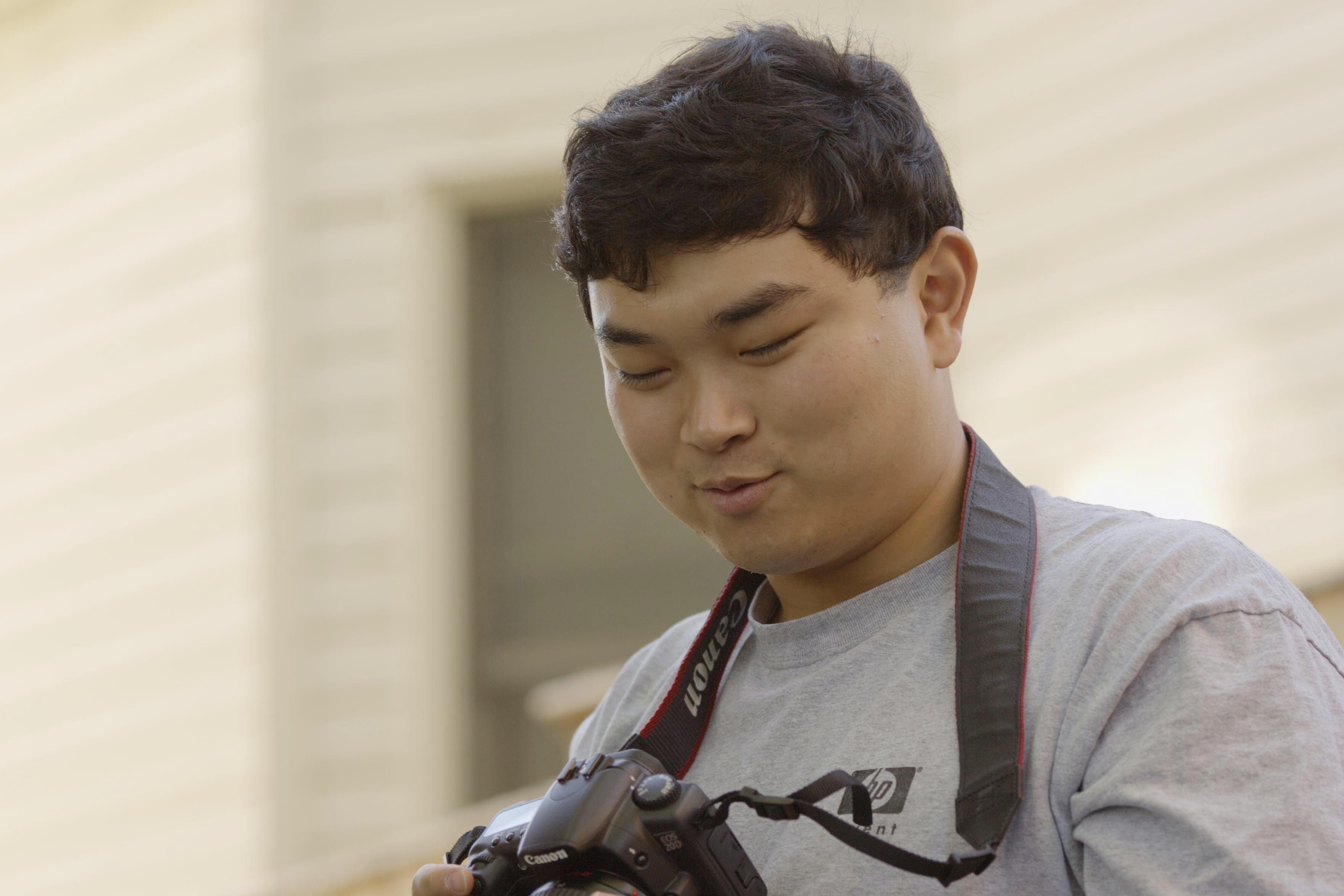}
        \includegraphics[width=\textwidth]{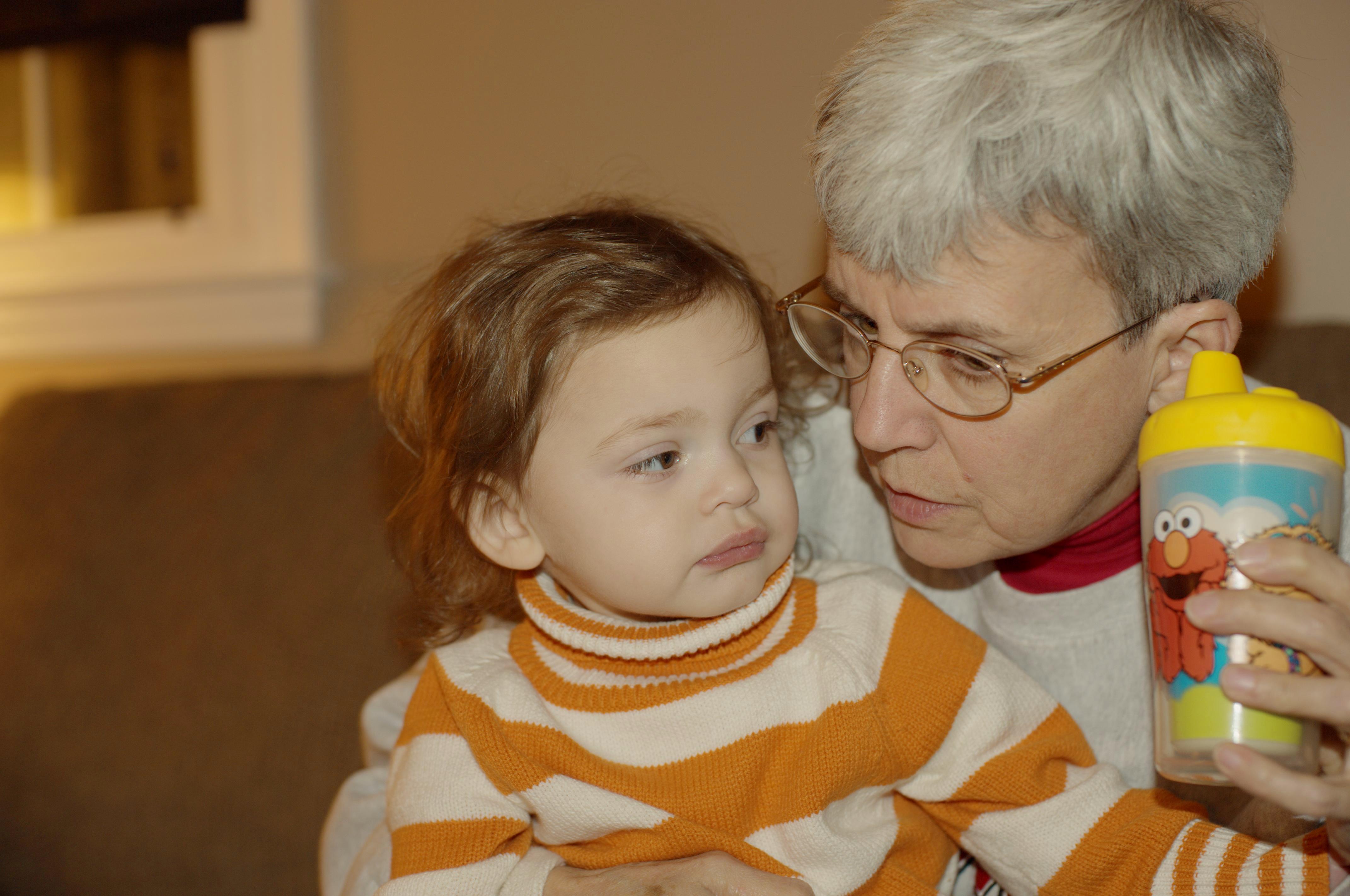} 
        \includegraphics[width=\textwidth]{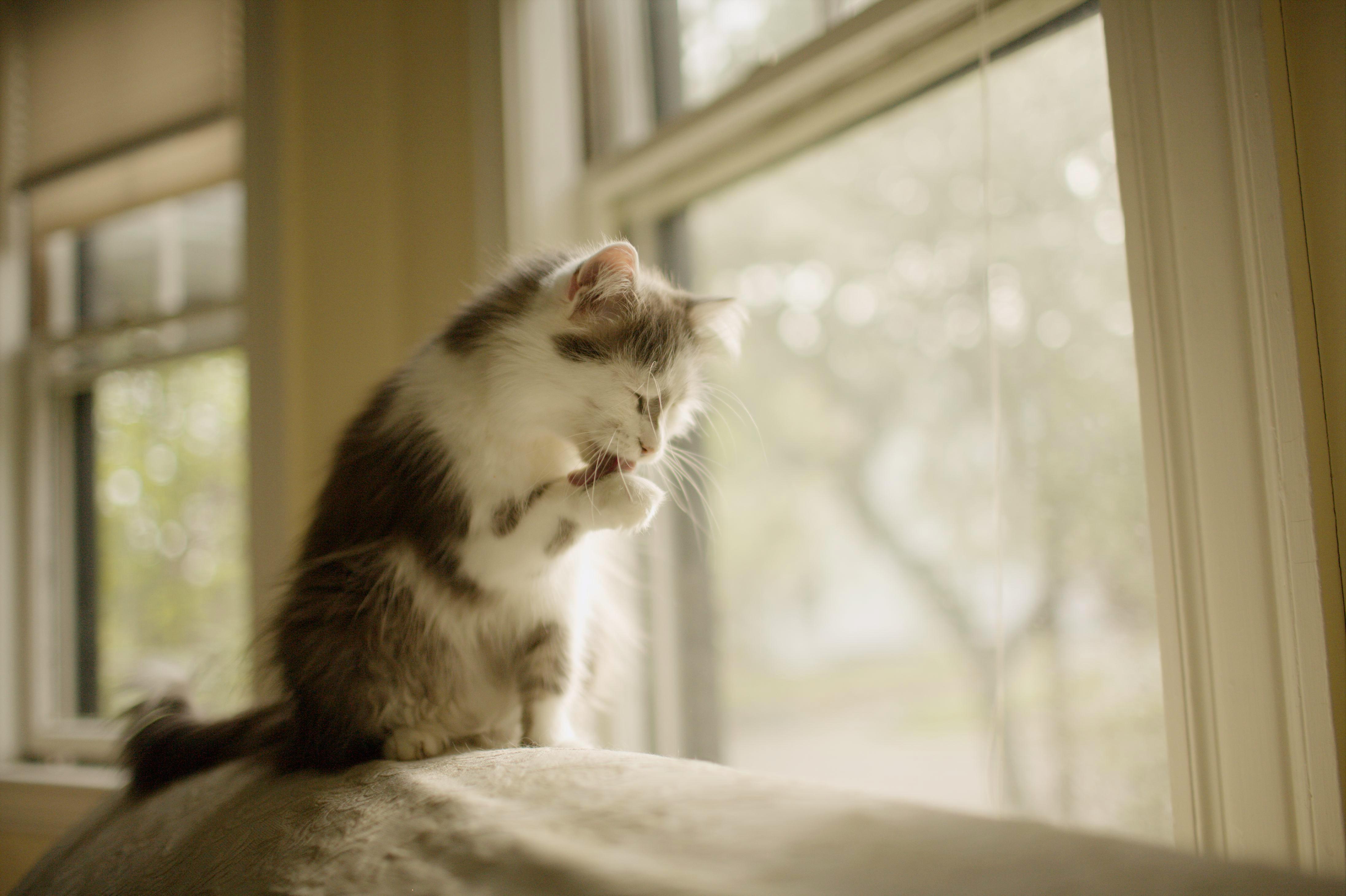}
        \includegraphics[width=\textwidth]{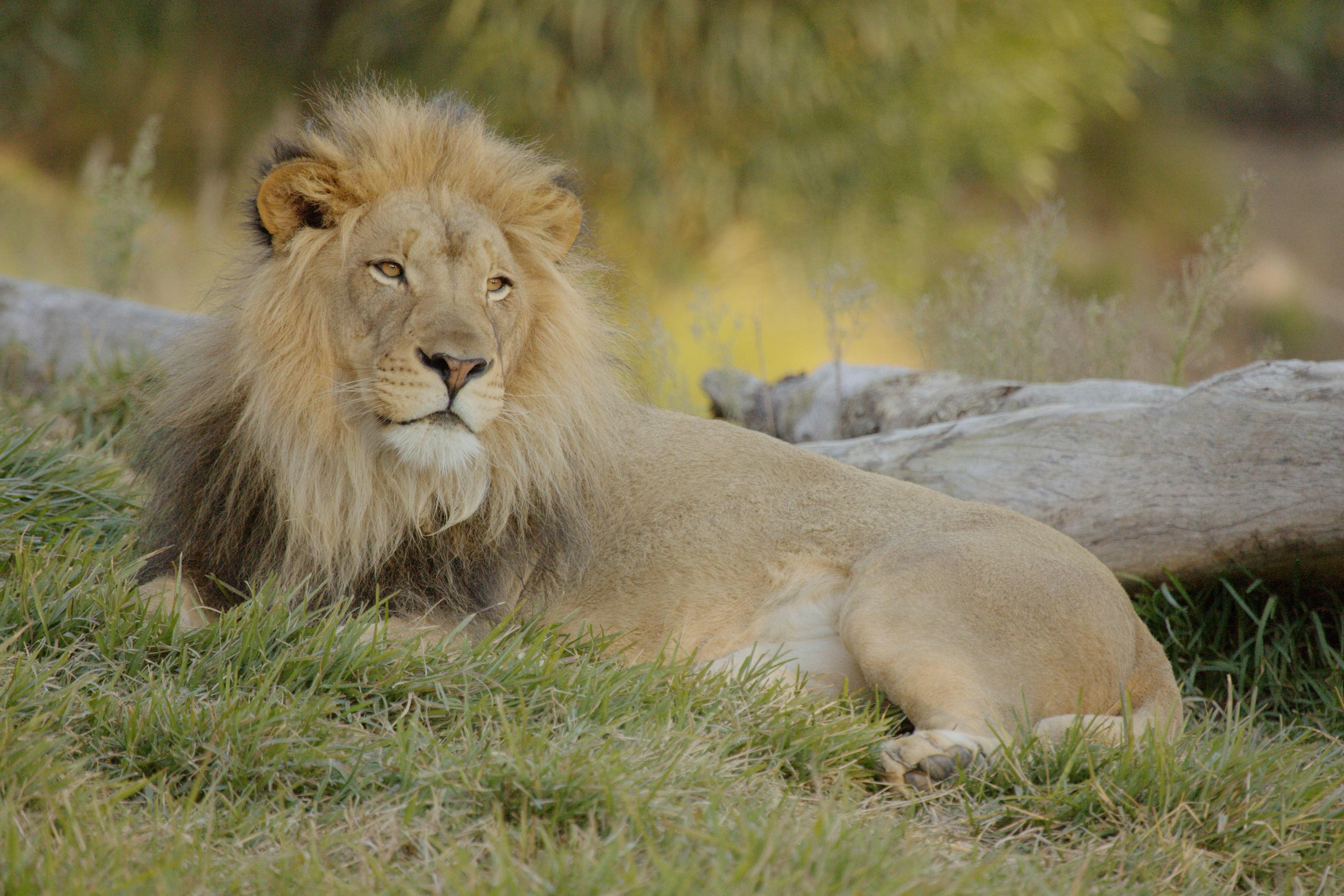} 
        \caption{DeNIM + Style WB}
    \end{subfigure}
    \caption{Comparison of the qualitative results of our efficient AWB correction method, namely \textit{DeNIM}, with the prior works on the selected samples from MIT-Adobe FiveK dataset \cite{global_tonal_adj}. We compare our results with Mixed WB \cite{Afifi_2022_WACV} and Style WB \cite{Kinli_2023_WACV}. Image indices from top to bottom: $323$, $606$, $2431$, $2808$, $2838$.}\label{fig:qual-mit} 
\end{figure*}

For qualitative and quantitative evaluation scenarios, we have used three different evaluation sets: Cube+ \cite{banic2019unsupervised} and MIT-Adobe FiveK \cite{global_tonal_adj}, along with the night photography rendering set \cite{Shutova_2023_CVPR}. The Cube+ dataset consists of 1,707 single illumination color-calibrated images, captured with a Canon EOS 550D camera during various seasons. The MIT-Adobe FiveK dataset comprises 5,000 images captured by different DSLR cameras, with each image manually retouched by multiple experts to correct the white balance. 

\section{Results and Discussion}

This section presents a detailed review of notable findings in our experiments. We primarily focus on three aspects while analyzing the results obtained in our experiments: visual quality, numeric evaluation, and efficiency. Qualitative analysis is conducted by comparing the results obtained by Mixed WB \cite{Afifi_2022_WACV}, Style WB \cite{Kinli_2023_WACV} and our strategy built on top of both methods on MIT-Adobe FiveK dataset and night photography rendering set. Following the literature, the evaluation of performance using quantitative metrics, analysis of model complexity, and comparison of efficiencies are all conducted using the Cube+ dataset.

\begin{table*}[ht]
\centering
\caption{Benchmark on single-illuminant Cube+ dataset \cite{banic2019unsupervised}. Following the prior works \cite{Afifi_2022_WACV, Kinli_2023_WACV}, we reported the mean, first (\textbf{Q1}), second (\textbf{Q2}) and third (\textbf{Q3}) quantile of mean-squared error (\textbf{MSE}), mean angular error (\textbf{MAE}) and color difference ($\Delta$\textbf{E 2000}) metrics. Different WB settings are denoted as \texttt{\{t,f,d,c,s\}}, which refers to tungsten, fluorescent, daylight, cloudy, and shade, respectively. $p$ refers to the patch size. The top results are indicated with colored cells as, the best: {\color{green}{\textbf{green}}}, the second: {\color{yellow}{yellow}}, the third: {\color{red}{red}}.}
\resizebox{\textwidth}{!}{
\begin{tabular}{|lccccccccccccc|}
\hline
\multicolumn{1}{|c|}{\multirow{2}{*}{\textbf{Method}}}                                                       & \multicolumn{4}{c|}{\textbf{MSE}}                                                                                                           & \multicolumn{4}{c|}{\textbf{MAE}}                                                                                                           & \multicolumn{4}{c|}{$\Delta$\textbf{E 2000}}   & \multicolumn{1}{|c|}{\multirow{2}{*}{\textbf{Size}}}                                                                                   \\ \cline{2-13} 
\multicolumn{1}{|c|}{}                                                                                       & \multicolumn{1}{c|}{\textbf{Mean}} & \multicolumn{1}{c|}{\textbf{Q1}} & \multicolumn{1}{c|}{\textbf{Q2}} & \multicolumn{1}{c|}{\textbf{Q3}} & \multicolumn{1}{c|}{\textbf{Mean}} & \multicolumn{1}{c|}{\textbf{Q1}} & \multicolumn{1}{c|}{\textbf{Q2}} & \multicolumn{1}{c|}{\textbf{Q3}} & \multicolumn{1}{c|}{\textbf{Mean}} & \multicolumn{1}{c|}{\textbf{Q1}} & \multicolumn{1}{c|}{\textbf{Q2}} & \multicolumn{1}{c|}{\textbf{Q3}} & \\ \hline
\multicolumn{1}{|l|}{FC4 \cite{Hu_2017_CVPR}}                                                                      & \multicolumn{1}{c|}{371.90}          & \multicolumn{1}{c|}{79.15}        & \multicolumn{1}{c|}{213.41}        & \multicolumn{1}{c|}{467.33}        & \multicolumn{1}{c|}{6.49$^{\circ}$}          & \multicolumn{1}{c|}{3.34$^{\circ}$}        & \multicolumn{1}{c|}{5.59$^{\circ}$}        & \multicolumn{1}{c|}{8.59$^{\circ}$}        & \multicolumn{1}{c|}{10.38}          & \multicolumn{1}{c|}{6.60}        & \multicolumn{1}{c|}{9.76}        & \multicolumn{1}{c|}{13.26}     
   & 5.89 MB     \\ \hline
\multicolumn{1}{|l|}{Quasi-U CC \cite{Bianco_2019_CVPR}}                                                                      & \multicolumn{1}{c|}{292.18}          & \multicolumn{1}{c|}{15.57}        & \multicolumn{1}{c|}{55.41}        & \multicolumn{1}{c|}{261.58}        & \multicolumn{1}{c|}{6.12$^{\circ}$}          & \multicolumn{1}{c|}{1.95$^{\circ}$}        & \multicolumn{1}{c|}{3.88$^{\circ}$}        & \multicolumn{1}{c|}{8.83$^{\circ}$}        & \multicolumn{1}{c|}{7.25}          & \multicolumn{1}{c|}{2.89}        & \multicolumn{1}{c|}{5.21}        & \multicolumn{1}{c|}{10.37}    &  622 MB  \\ \hline
\multicolumn{1}{|l|}{KNN WB \cite{Afifi2019WBsRGBImages}}                                                                      & \multicolumn{1}{c|}{194.98}          & \multicolumn{1}{c|}{27.43}        & \multicolumn{1}{c|}{57.08}        & \multicolumn{1}{c|}{118.21}        & \multicolumn{1}{c|}{4.12$^{\circ}$}          & \multicolumn{1}{c|}{1.96$^{\circ}$}        & \multicolumn{1}{c|}{3.17$^{\circ}$}        & \multicolumn{1}{c|}{5.04$^{\circ}$}        & \multicolumn{1}{c|}{5.68}          & \multicolumn{1}{c|}{3.22}        & \multicolumn{1}{c|}{4.61}        & \multicolumn{1}{c|}{6.70}       &   21.8 MB  \\ \hline
\multicolumn{1}{|l|}{Interactive WB \cite{afifi2020interactive}}                                                                & \multicolumn{1}{c|}{159.88}          & \multicolumn{1}{c|}{21.94}        & \multicolumn{1}{c|}{54.76}        & \multicolumn{1}{c|}{125.02}        & \multicolumn{1}{c|}{4.64$^{\circ}$}          & \multicolumn{1}{c|}{2.12$^{\circ}$}        & \multicolumn{1}{c|}{3.64$^{\circ}$}        & \multicolumn{1}{c|}{5.98$^{\circ}$}        & \multicolumn{1}{c|}{6.20}          & \multicolumn{1}{c|}{3.28}        & \multicolumn{1}{c|}{5.17}        & \multicolumn{1}{c|}{7.45}    &  \cellcolor[HTML]{C0FDBF}\textbf{38 KB}  \\ \hline
\multicolumn{1}{|l|}{Deep WB \cite{Afifi_2020_CVPR}}                                                                      & \multicolumn{1}{c|}{\cellcolor[HTML]{FFFFC7}80.46}          & \multicolumn{1}{c|}{15.43}        & \multicolumn{1}{c|}{33.88}        & \multicolumn{1}{c|}{74.42}        & \multicolumn{1}{c|}{3.45$^{\circ}$}          & \multicolumn{1}{c|}{1.87$^{\circ}$}        & \multicolumn{1}{c|}{2.82$^{\circ}$}        & \multicolumn{1}{c|}{4.26$^{\circ}$}        & \multicolumn{1}{c|}{4.59}          & \multicolumn{1}{c|}{2.68}        & \multicolumn{1}{c|}{3.81}        & \multicolumn{1}{c|}{5.53}     &   16.7 MB     \\ \hline
\multicolumn{1}{|l|}{MIMT \cite{li2023mimt}}  & \multicolumn{1}{c|}{-}        & \multicolumn{1}{c|}{-}       & \multicolumn{1}{c|}{-}       & \multicolumn{1}{c|}{-}       & \multicolumn{1}{c|}{2.52$^{\circ}$}          & \multicolumn{1}{c|}{0.98$^{\circ}$}        & \multicolumn{1}{c|}{1.38$^{\circ}$}        & \multicolumn{1}{c|}{2.96$^{\circ}$}        & \multicolumn{1}{c|}{\cellcolor[HTML]{FFFFC7}2.88}          & \multicolumn{1}{c|}{1.94}        & \multicolumn{1}{c|}{2.42}        & \multicolumn{1}{c|}{\cellcolor[HTML]{FFFFC7}2.87}    &   - \\ \hline

\multicolumn{14}{|c|}{\cellcolor[HTML]{B4EBF1}\textbf{Mixed WB \cite{Afifi_2022_WACV}}}   \\ \hline
\multicolumn{1}{|l|}{$p = 64$, WB=\texttt{\{t,d,s\}}}                                                                      & \multicolumn{1}{c|}{168.38}          & \multicolumn{1}{c|}{\cellcolor[HTML]{FFCCC9}8.97}        & \multicolumn{1}{c|}{19.87}        & \multicolumn{1}{c|}{105.22}        & \multicolumn{1}{c|}{4.20$^{\circ}$}          & \multicolumn{1}{c|}{1.39$^{\circ}$}        & \multicolumn{1}{c|}{2.18$^{\circ}$}        & \multicolumn{1}{c|}{5.54$^{\circ}$}        & \multicolumn{1}{c|}{5.03}          & \multicolumn{1}{c|}{2.07}        & \multicolumn{1}{c|}{3.12}        & \multicolumn{1}{c|}{7.19}     &    \cellcolor[HTML]{FFFFC7}5.09 MB    \\ \hline
\multicolumn{1}{|l|}{$p = 64$, WB=\texttt{\{t,f,d,c,s\}}}                                                                      & \multicolumn{1}{c|}{161.80}          & \multicolumn{1}{c|}{9.01}        & \multicolumn{1}{c|}{19.33}        & \multicolumn{1}{c|}{90.81}        & \multicolumn{1}{c|}{4.05$^{\circ}$}          & \multicolumn{1}{c|}{1.40$^{\circ}$}        & \multicolumn{1}{c|}{2.12$^{\circ}$}        & \multicolumn{1}{c|}{4.88$^{\circ}$}        & \multicolumn{1}{c|}{4.89}          & \multicolumn{1}{c|}{2.16}        & \multicolumn{1}{c|}{3.10}        & \multicolumn{1}{c|}{6.78}     &    \cellcolor[HTML]{FFCCC9}5.10 MB    \\ \hline
\multicolumn{1}{|l|}{$p = 128$, WB=\texttt{\{t,f,d,c,s\}}}                                                                      & \multicolumn{1}{c|}{176.38}          & \multicolumn{1}{c|}{16.96}        & \multicolumn{1}{c|}{35.91}        & \multicolumn{1}{c|}{115.50}        & \multicolumn{1}{c|}{4.71$^{\circ}$}          & \multicolumn{1}{c|}{2.10$^{\circ}$}        & \multicolumn{1}{c|}{3.09$^{\circ}$}        & \multicolumn{1}{c|}{5.92$^{\circ}$}        & \multicolumn{1}{c|}{5.77}          & \multicolumn{1}{c|}{3.01}        & \multicolumn{1}{c|}{4.27}        & \multicolumn{1}{c|}{7.71}     &   \cellcolor[HTML]{FFCCC9}5.10 MB    \\ \hline

\multicolumn{14}{|c|}{\cellcolor[HTML]{B4EBF1}\textbf{Style WB \cite{Kinli_2023_WACV}}}   \\ \hline
\multicolumn{1}{|l|}{$p = 64$, WB=\texttt{\{t,d,s\}}}         & \multicolumn{1}{c|}{92.65}         & \multicolumn{1}{c|}{\cellcolor[HTML]{C0FDBF}\textbf{6.52}}       & \multicolumn{1}{c|}{\cellcolor[HTML]{C0FDBF}\textbf{14.23}}       & \multicolumn{1}{c|}{\cellcolor[HTML]{FFFFC7}35.01}      & \multicolumn{1}{c|}{2.47$^{\circ}$}         & \multicolumn{1}{c|}{\cellcolor[HTML]{FFFFC7}0.82$^{\circ}$}        & \multicolumn{1}{c|}{1.44$^{\circ}$}        & \multicolumn{1}{c|}{2.49$^{\circ}$}        & \multicolumn{1}{c|}{2.99}          & \multicolumn{1}{c|}{\cellcolor[HTML]{C0FDBF}\textbf{1.36}}       & \multicolumn{1}{c|}{\cellcolor[HTML]{FFFFC7}2.04}        & \multicolumn{1}{c|}{3.32}   &    61.0 MB    \\ \hline

\multicolumn{1}{|l|}{$p = 64$, WB=\texttt{\{t,f,d,c,s\}}}   & \multicolumn{1}{c|}{151.38}        & \multicolumn{1}{c|}{29.49}       & \multicolumn{1}{c|}{56.35}       & \multicolumn{1}{c|}{125.33}      & \multicolumn{1}{c|}{4.18$^{\circ}$}          & \multicolumn{1}{c|}{2.13$^{\circ}$}        & \multicolumn{1}{c|}{3.03$^{\circ}$}        & \multicolumn{1}{c|}{4.81$^{\circ}$}        & \multicolumn{1}{c|}{5.42}          & \multicolumn{1}{c|}{3.11}        & \multicolumn{1}{c|}{4.42}        & \multicolumn{1}{c|}{6.76}      &  61.1 MB \\ \hline

\multicolumn{1}{|l|}{$p = 128$, WB=\texttt{\{t,d,s\}}}  & \multicolumn{1}{c|}{88.03}        & \multicolumn{1}{c|}{\cellcolor[HTML]{FFFFC7}7.92}       & \multicolumn{1}{c|}{\cellcolor[HTML]{FFCCC9}17.73}       & \multicolumn{1}{c|}{45.01}       & \multicolumn{1}{c|}{2.61$^{\circ}$}          & \multicolumn{1}{c|}{0.93$^{\circ}$}        & \multicolumn{1}{c|}{1.58$^{\circ}$}        & \multicolumn{1}{c|}{2.85$^{\circ}$}        & \multicolumn{1}{c|}{3.24}          & \multicolumn{1}{c|}{\cellcolor[HTML]{FFFFC7}1.50}        & \multicolumn{1}{c|}{\cellcolor[HTML]{FFCCC9}2.30}        & \multicolumn{1}{c|}{3.95}     &  61.2 MB \\ \hline

\multicolumn{1}{|l|}{$p = 128$, WB=\texttt{\{t,f,d,c,s\}}}  & \multicolumn{1}{c|}{100.24}        & \multicolumn{1}{c|}{10.77}       & \multicolumn{1}{c|}{37.74}       & \multicolumn{1}{c|}{70.18}       & \multicolumn{1}{c|}{3.09$^{\circ}$}          & \multicolumn{1}{c|}{1.15$^{\circ}$}        & \multicolumn{1}{c|}{2.61$^{\circ}$}        & \multicolumn{1}{c|}{3.87$^{\circ}$}        & \multicolumn{1}{c|}{3.96}          & \multicolumn{1}{c|}{\cellcolor[HTML]{FFCCC9}1.59}        & \multicolumn{1}{c|}{3.55}        & \multicolumn{1}{c|}{5.51}    &   61.3 MB \\ \hline

\multicolumn{14}{|c|}{\cellcolor[HTML]{B4EBF1}\textbf{DeNIM + Mixed WB \cite{Afifi_2022_WACV} (ours)}}   \\ \hline
\multicolumn{1}{|l|}{$p = 64$, WB=\texttt{\{t,d,s\}}}         & \multicolumn{1}{c|}{120.14}         & \multicolumn{1}{c|}{36.39}        & \multicolumn{1}{c|}{77.40}       & \multicolumn{1}{c|}{152.96}       & \multicolumn{1}{c|}{2.57$^{\circ}$}          & \multicolumn{1}{c|}{1.53$^{\circ}$}        & \multicolumn{1}{c|}{2.17$^{\circ}$}       & \multicolumn{1}{c|}{3.19$^{\circ}$}       & \multicolumn{1}{c|}{5.26}          & \multicolumn{1}{c|}{3.38}       & \multicolumn{1}{c|}{4.71}        & \multicolumn{1}{c|}{6.64}   &    28.7 MB    \\ \hline

\multicolumn{1}{|l|}{$p = 64$, WB=\texttt{\{t,f,d,c,s\}}}         & \multicolumn{1}{c|}{129.01}         & \multicolumn{1}{c|}{14.39}        & \multicolumn{1}{c|}{27.69}       & \multicolumn{1}{c|}{57.90}       & \multicolumn{1}{c|}{2.67$^{\circ}$}          & \multicolumn{1}{c|}{0.99$^{\circ}$}        & \multicolumn{1}{c|}{1.45$^{\circ}$}       & \multicolumn{1}{c|}{2.29$^{\circ}$}       & \multicolumn{1}{c|}{3.96}          & \multicolumn{1}{c|}{2.10}       & \multicolumn{1}{c|}{2.85}        & \multicolumn{1}{c|}{4.24}   &    28.7 MB    \\ \hline
				
\multicolumn{1}{|l|}{$p = 128$, WB=\texttt{\{t,d,s\}}}         & \multicolumn{1}{c|}{158.58}         & \multicolumn{1}{c|}{60.14}        & \multicolumn{1}{c|}{115.66}       & \multicolumn{1}{c|}{198.59}       & \multicolumn{1}{c|}{4.20$^{\circ}$}          & \multicolumn{1}{c|}{2.38$^{\circ}$}        & \multicolumn{1}{c|}{3.77$^{\circ}$}       & \multicolumn{1}{c|}{5.63$^{\circ}$}       & \multicolumn{1}{c|}{5.69}          & \multicolumn{1}{c|}{3.91}       & \multicolumn{1}{c|}{5.41}        & \multicolumn{1}{c|}{7.10}   &    28.8 MB    \\ \hline

\multicolumn{1}{|l|}{$p = 128$, WB=\texttt{\{t,f,d,c,s\}}}         & \multicolumn{1}{c|}{99.70}         & \multicolumn{1}{c|}{13.89}        & \multicolumn{1}{c|}{24.71}       & \multicolumn{1}{c|}{43.88}       & \multicolumn{1}{c|}{2.49$^{\circ}$}          & \multicolumn{1}{c|}{1.07$^{\circ}$}        & \multicolumn{1}{c|}{1.62$^{\circ}$}       & \multicolumn{1}{c|}{2.41$^{\circ}$}       & \multicolumn{1}{c|}{3.44}          & \multicolumn{1}{c|}{1.95}       & \multicolumn{1}{c|}{2.74}        & \multicolumn{1}{c|}{3.78}   &    28.8 MB    \\ \hline

\multicolumn{14}{|c|}{\cellcolor[HTML]{B4EBF1}\textbf{DeNIM + Style WB \cite{Kinli_2023_WACV} (ours)}}   \\ \hline
\multicolumn{1}{|l|}{$p = 64$, WB=\texttt{\{t,d,s\}}}         & \multicolumn{1}{c|}{\cellcolor[HTML]{C0FDBF}\textbf{65.80}}         & \multicolumn{1}{c|}{10.06}        & \multicolumn{1}{c|}{\cellcolor[HTML]{FFFFC7}16.98}       & \multicolumn{1}{c|}{\cellcolor[HTML]{C0FDBF}\textbf{28.82}}       & \multicolumn{1}{c|}{\cellcolor[HTML]{FFFFC7}2.03$^{\circ}$}          & \multicolumn{1}{c|}{\cellcolor[HTML]{FFCCC9}0.88$^{\circ}$}        & \multicolumn{1}{c|}{\cellcolor[HTML]{FFFFC7}1.23$^{\circ}$}       & \multicolumn{1}{c|}{\cellcolor[HTML]{FFFFC7}1.93$^{\circ}$}       & \multicolumn{1}{c|}{\cellcolor[HTML]{FFCCC9}2.95}          & \multicolumn{1}{c|}{1.79}       & \multicolumn{1}{c|}{2.33}        & \multicolumn{1}{c|}{\cellcolor[HTML]{FFCCC9}3.18}   &    196.3 MB    \\ \hline

\multicolumn{1}{|l|}{$p = 64$, WB=\texttt{\{t,f,d,c,s\}}}         & \multicolumn{1}{c|}{83.41}         & \multicolumn{1}{c|}{13.23}        & \multicolumn{1}{c|}{21.46}       & \multicolumn{1}{c|}{\cellcolor[HTML]{FFCCC9}37.44}       & \multicolumn{1}{c|}{\cellcolor[HTML]{C0FDBF}\textbf{1.93}$^{\circ}$}          & \multicolumn{1}{c|}{\cellcolor[HTML]{C0FDBF}\textbf{0.77}$^{\circ}$}        & \multicolumn{1}{c|}{\cellcolor[HTML]{C0FDBF}\textbf{1.09}$^{\circ}$}       & \multicolumn{1}{c|}{\cellcolor[HTML]{C0FDBF}\textbf{1.70}$^{\circ}$}       & \multicolumn{1}{c|}{\cellcolor[HTML]{C0FDBF}\textbf{2.73}}         & \multicolumn{1}{c|}{1.62}       & \multicolumn{1}{c|}{\cellcolor[HTML]{C0FDBF}\textbf{2.03}}        & \multicolumn{1}{c|}{\cellcolor[HTML]{C0FDBF}\textbf{2.71}}   &    196.3 MB    \\ \hline

\multicolumn{1}{|l|}{$p = 128$, WB=\texttt{\{t,d,s\}}}         & \multicolumn{1}{c|}{\cellcolor[HTML]{FFCCC9}80.53}         & \multicolumn{1}{c|}{17.59}        & \multicolumn{1}{c|}{27.80}       & \multicolumn{1}{c|}{44.35}       & \multicolumn{1}{c|}{\cellcolor[HTML]{FFCCC9}2.16$^{\circ}$}          & \multicolumn{1}{c|}{\cellcolor[HTML]{FFCCC9}0.88$^{\circ}$}        & \multicolumn{1}{c|}{\cellcolor[HTML]{FFCCC9}1.34$^{\circ}$}       & \multicolumn{1}{c|}{\cellcolor[HTML]{FFCCC9}2.16$^{\circ}$}       & \multicolumn{1}{c|}{3.08}          & \multicolumn{1}{c|}{1.86}       & \multicolumn{1}{c|}{2.37}        & \multicolumn{1}{c|}{3.30}   &    196.4 MB    \\ \hline

\multicolumn{1}{|l|}{$p = 128$, WB=\texttt{\{t,f,d,c,s\}}}         & \multicolumn{1}{c|}{89.10}         & \multicolumn{1}{c|}{11.27}        & \multicolumn{1}{c|}{19.34}       & \multicolumn{1}{c|}{43.01}       & \multicolumn{1}{c|}{2.49$^{\circ}$}          & \multicolumn{1}{c|}{1.24$^{\circ}$}        & \multicolumn{1}{c|}{1.64$^{\circ}$}       & \multicolumn{1}{c|}{2.92$^{\circ}$}       & \multicolumn{1}{c|}{3.16}          & \multicolumn{1}{c|}{1.87}       & \multicolumn{1}{c|}{2.53}        & \multicolumn{1}{c|}{3.35}   &    196.4 MB    \\ \hline

\end{tabular}}
\label{table:results_1}
\end{table*}

\begin{table}[]
\centering
\caption{Comparison of the complexity of our method and the prior methods with their post-processing tricks. $ms$: multi-scale weighting maps, $eas$: edge-aware smoothing.}
\begin{adjustbox}{width=0.48\textwidth}
\begin{tabular}{l|c|c|c}
\textbf{Model Architecture} & \multicolumn{1}{l|}{\textbf{Time (s)}} & \multicolumn{1}{l|}{\textbf{Param (M)}} & \multicolumn{1}{l}{\textbf{FLOPS (G)}} \\ \hline
Mixed WB \cite{Afifi_2022_WACV} + $ms$ + $eas$           & 10.390                                                    & \multirow{4}{*}{\textbf{1.32}}                   & \multirow{4}{*}{82.68}                 \\ \cline{1-2}
Mixed WB \cite{Afifi_2022_WACV} + $ms$                & 0.228                                                    &                                         &                                        \\ \cline{1-2}
Mixed WB \cite{Afifi_2022_WACV} + $eas$                & 10.279                                                   &                                         &                                        \\ \cline{1-2}
Mixed WB \cite{Afifi_2022_WACV}                     & 0.212                                                    &                                         &                                        \\ \hline
Style WB \cite{Kinli_2023_WACV} + $ms$ + $eas$           & 10.342                                                   & \multirow{4}{*}{15.31}                  & \multirow{4}{*}{76.80}                  \\ \cline{1-2}
Style WB \cite{Kinli_2023_WACV} + $ms$                & 0.232                                                    &                                         &                                        \\ \cline{1-2}
Style WB \cite{Kinli_2023_WACV} + $eas$                & 10.307                                                   &                                         &                                        \\ \cline{1-2}
Style WB \cite{Kinli_2023_WACV}                     & 0.217                                                    &                                         &                                        \\ \hline
Ours w/ Mixed WB \cite{Afifi_2022_WACV}    & \textbf{0.006}                                                    & 1.67                                    & \textbf{2.14}                                   \\ \hline
Ours w/ Style WB \cite{Kinli_2023_WACV}    & \textbf{0.010}                                                     & 16.19                                   & 26.89                                 

\end{tabular}
\end{adjustbox}
\label{table:results_2}
\end{table}

\noindent\textbf{Qualitative analysis:} To use the images in MIT-Adobe FiveK dataset for our experiments, we first render the linear raw DNG images with different WB settings (\textit{e.g.,} Daylight, Tungsten, Shade) by using the method presented in \cite{afifi2019color}. Figure \ref{fig:qual-mit} demonstrates the qualitative comparison of our AWB correction results and the prior works' on selected samples from the dataset. The indices of selected samples in the dataset are as follows: $323$, $606$, $2431$, $2808$, and $2838$. These results indicate that our proposed strategy performs comparably well to the prior works on a per-pixel basis for AWB correction in the sRGB space. Utilizing per-pixel color mapping seems to result in color casts that are closer to human perception by more accurately representing the lighting conditions within the scene.

\begin{figure*}[ht!]
    \centering
    \begin{subfigure}{0.24\textwidth}
        \includegraphics[width=\textwidth]{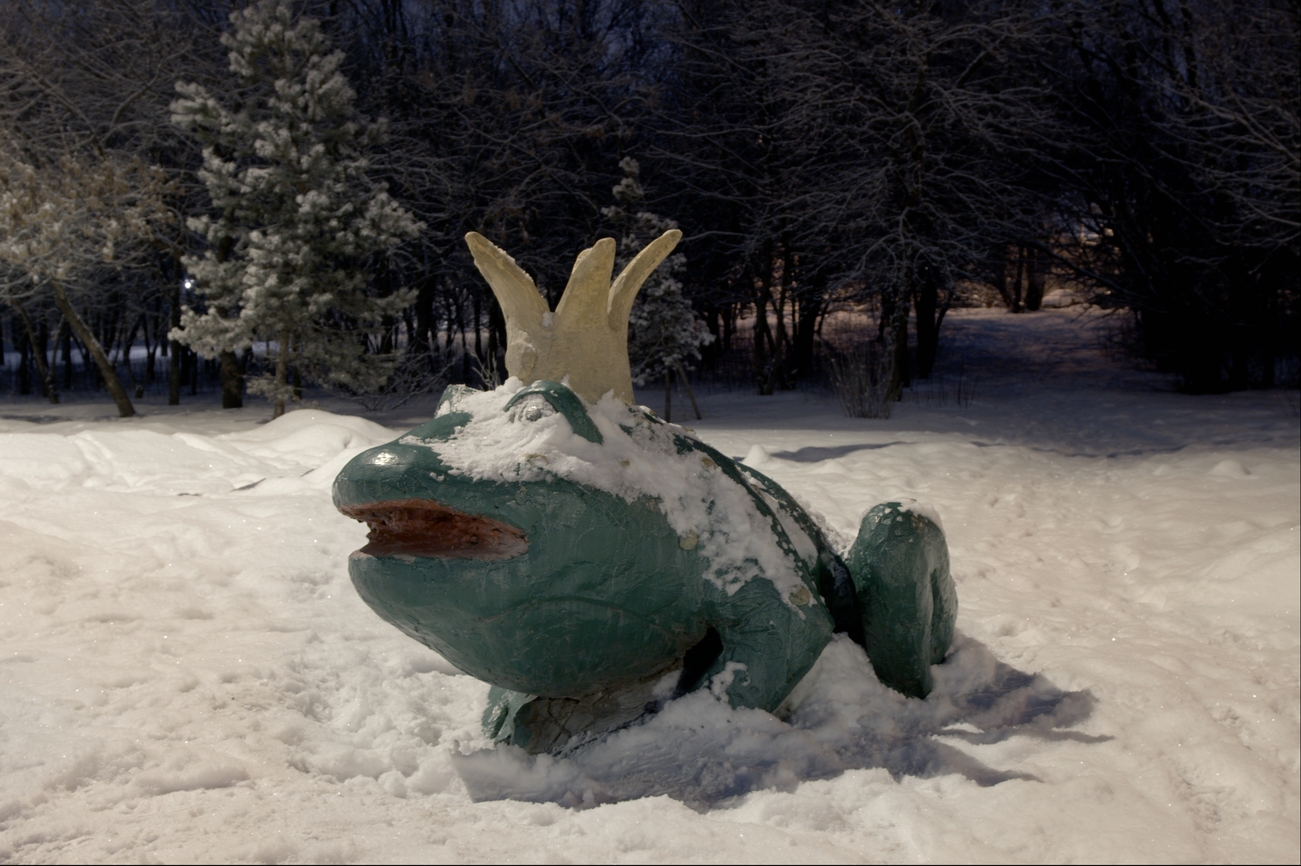} 
        \includegraphics[width=\textwidth]{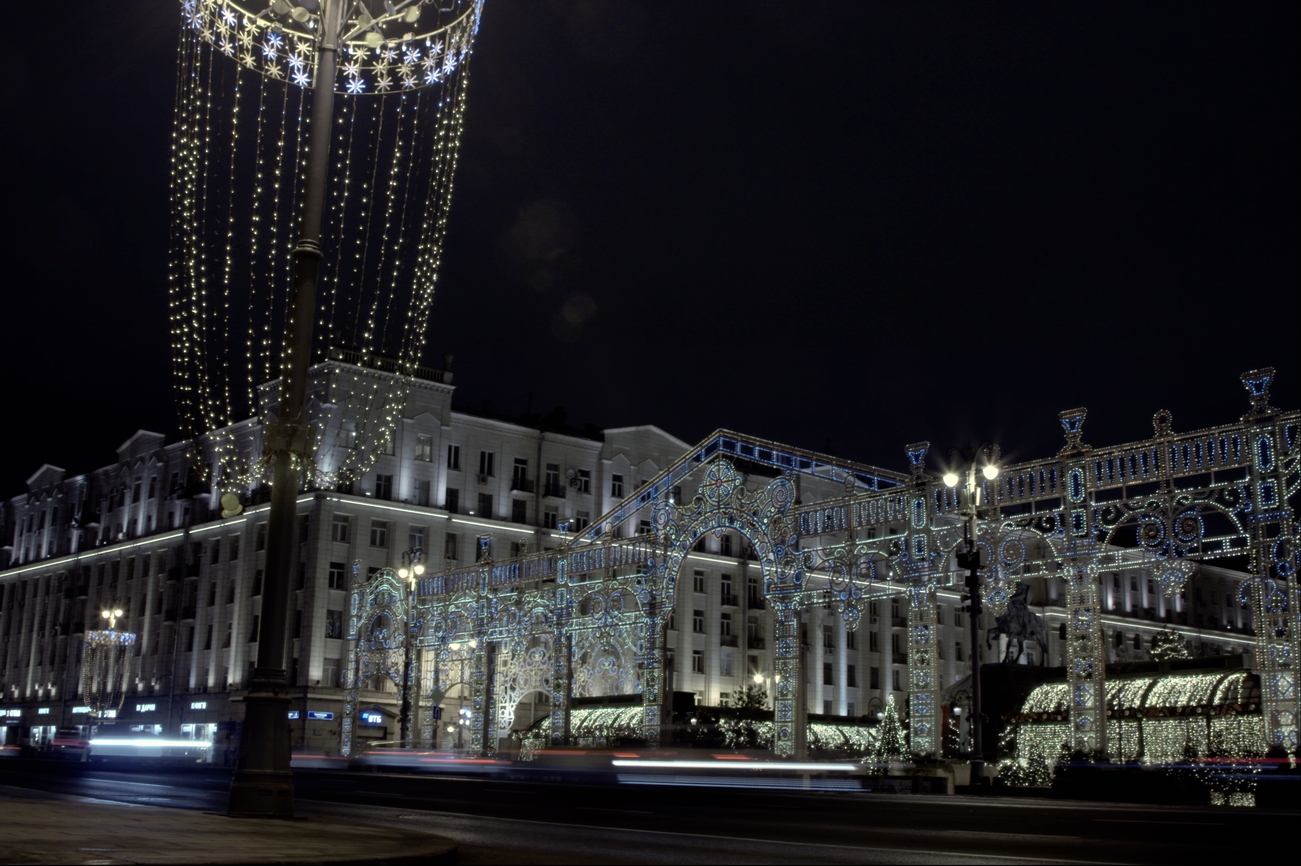}
        \includegraphics[width=\textwidth]{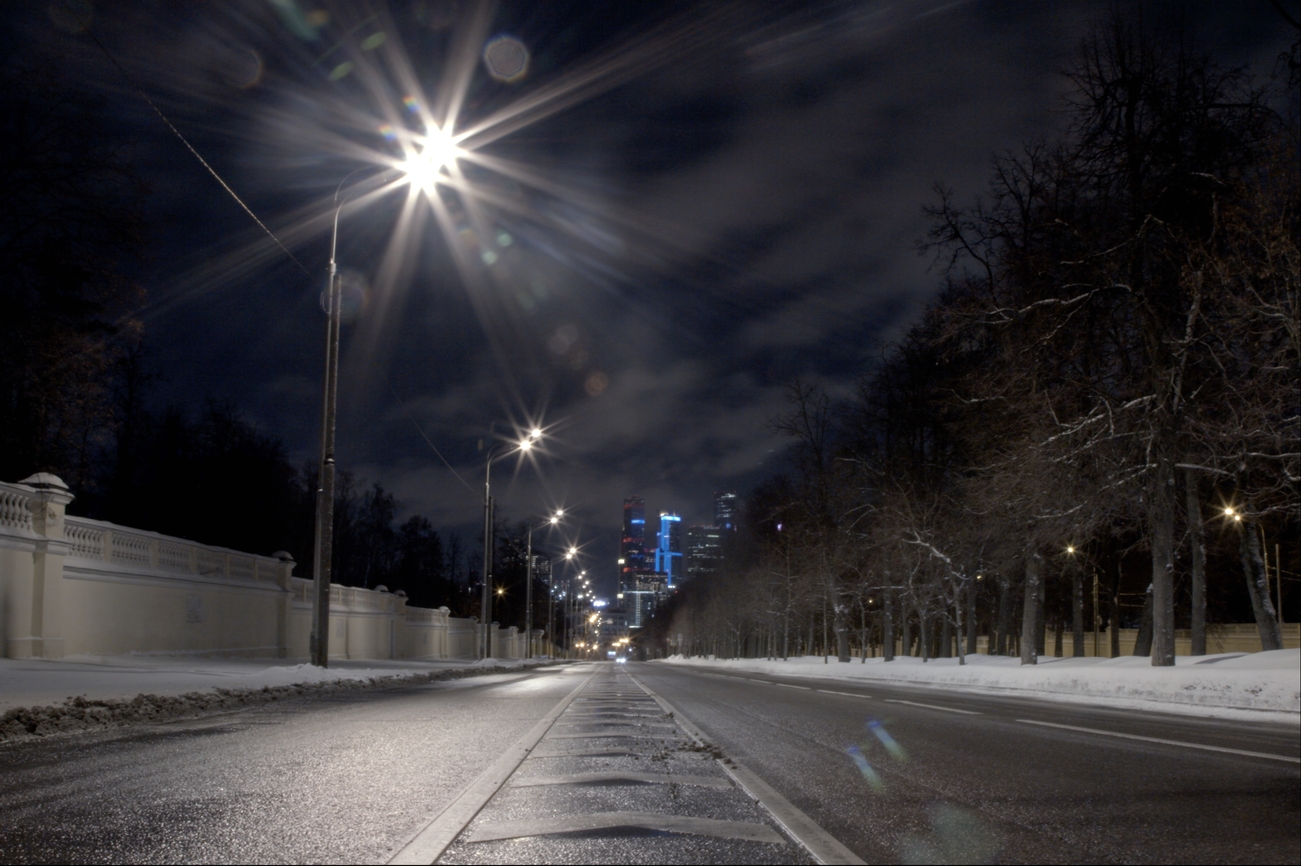}
        \includegraphics[width=\textwidth]{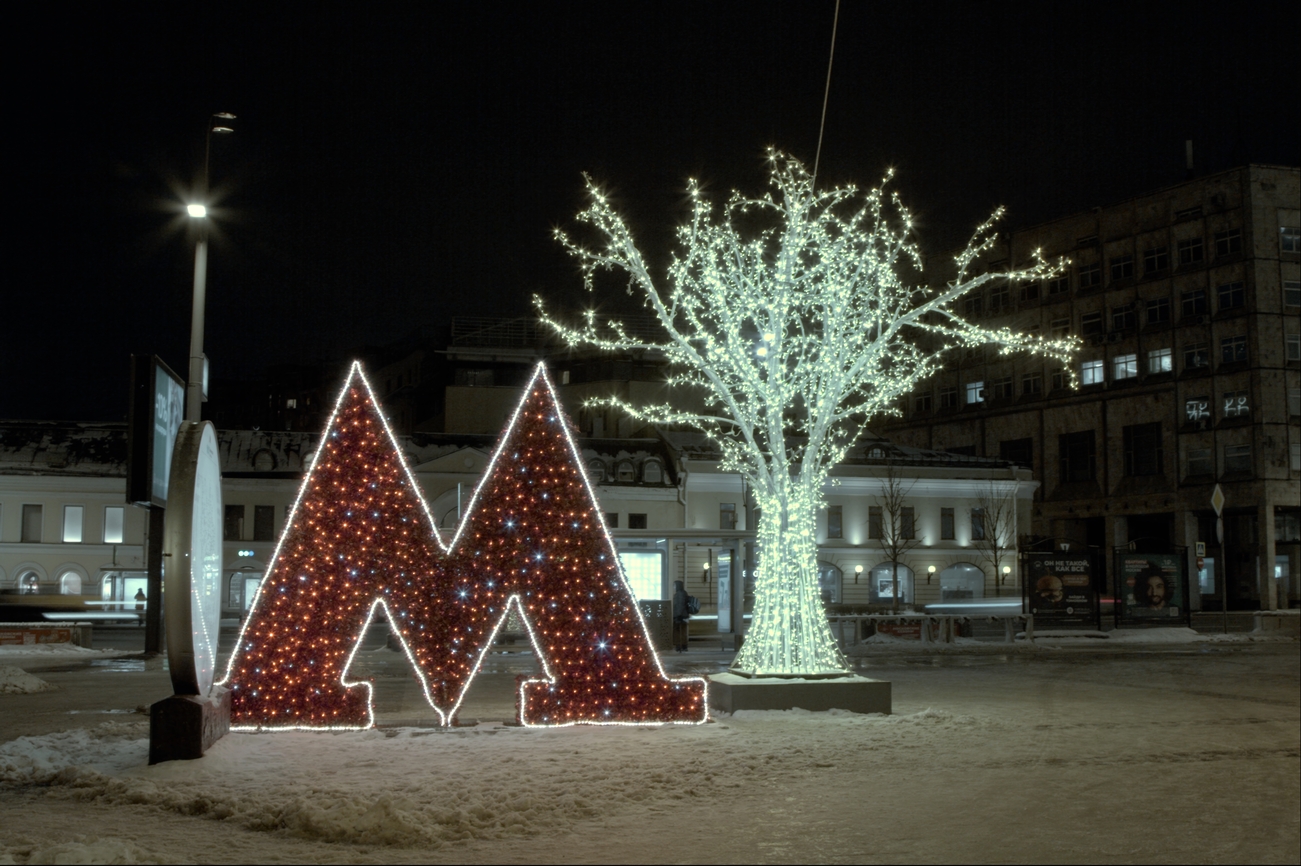}
        \includegraphics[width=\textwidth]{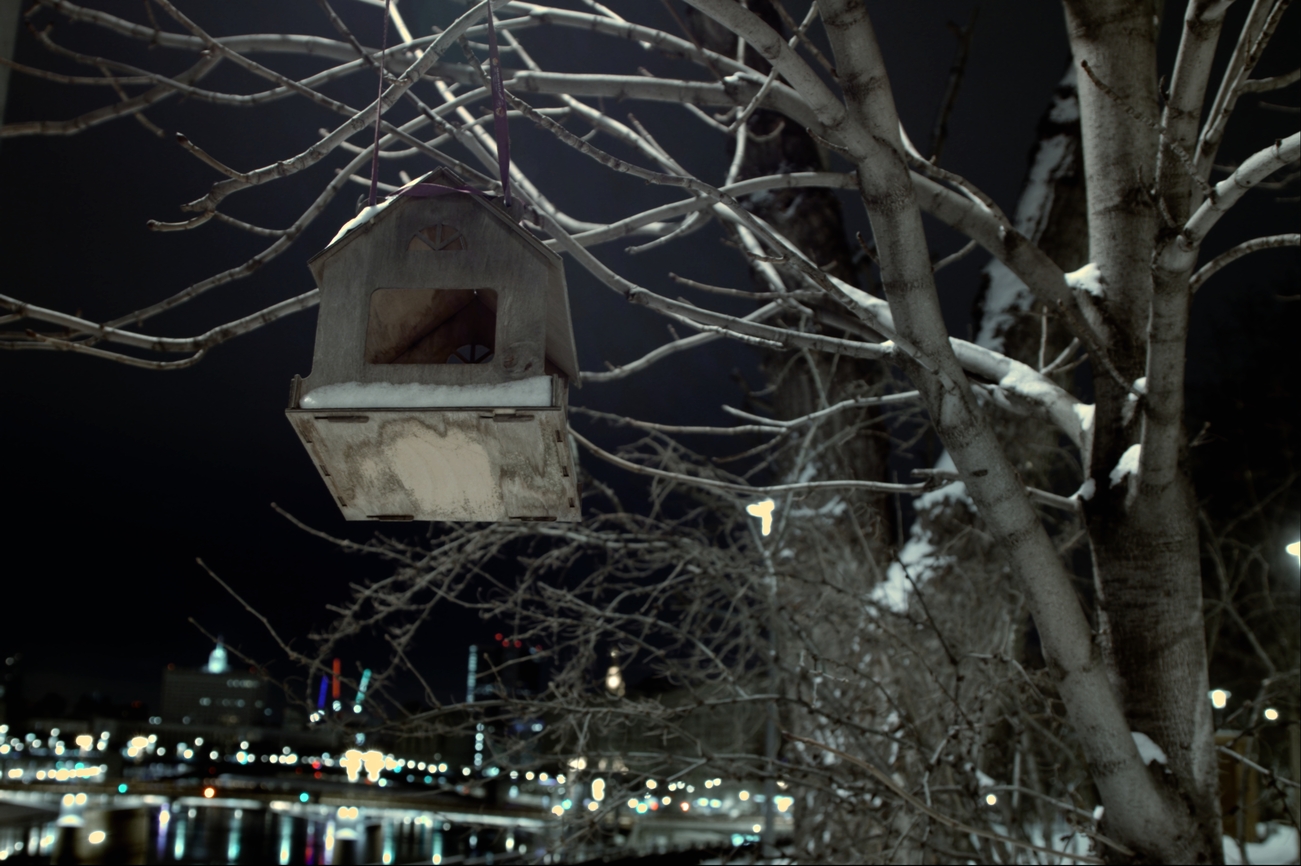}
        \caption{Mixed WB}
    \end{subfigure}
    \begin{subfigure}{0.24\textwidth}
        \includegraphics[width=\textwidth]{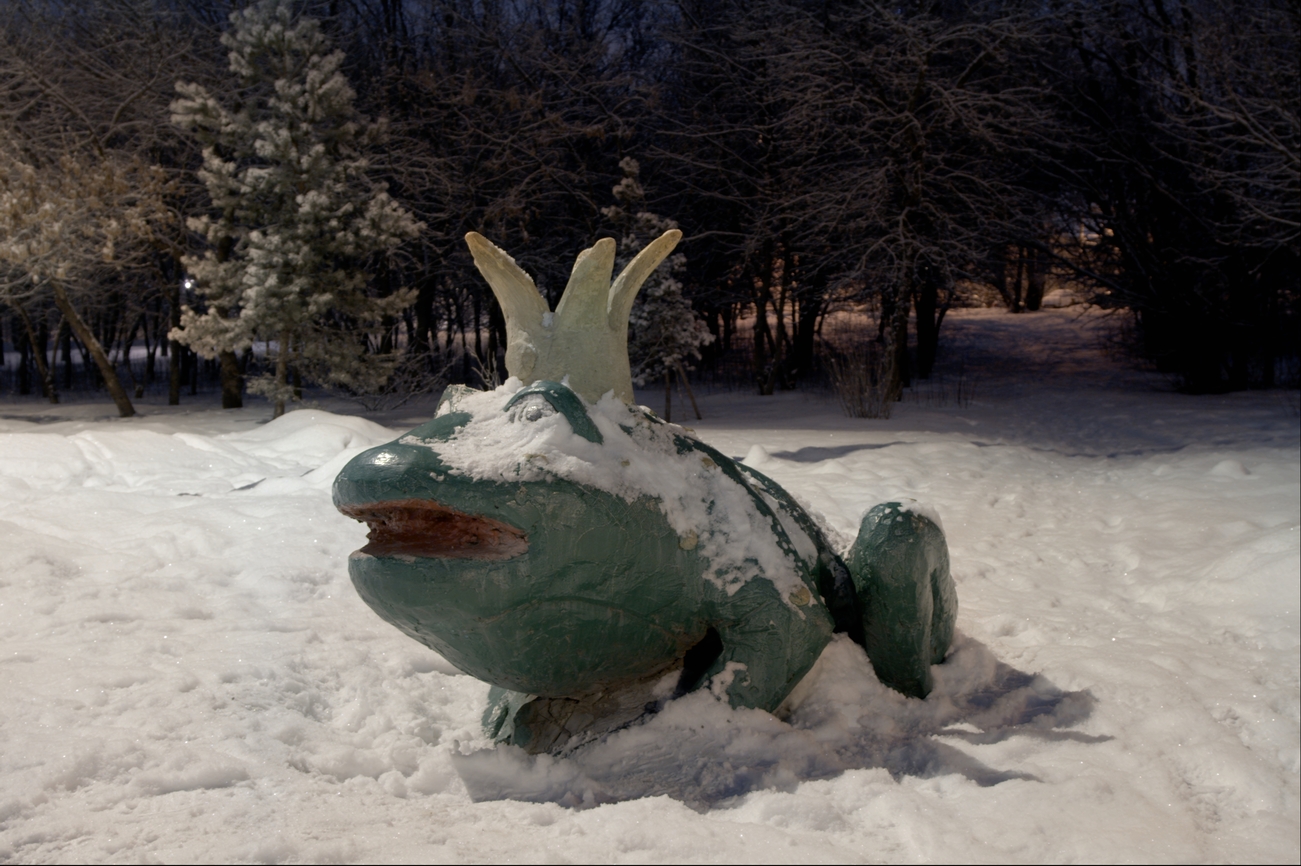} 
        \includegraphics[width=\textwidth]{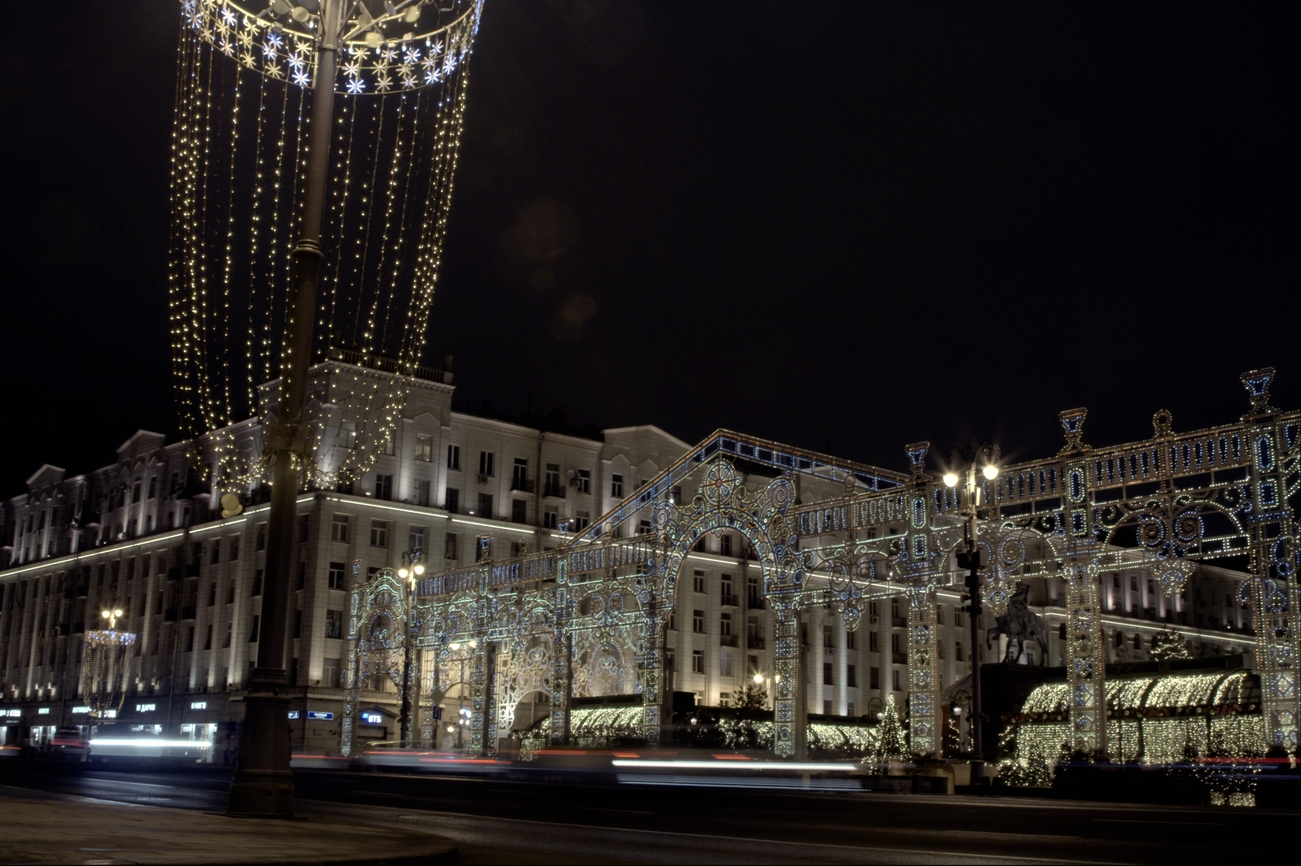}
        \includegraphics[width=\textwidth]{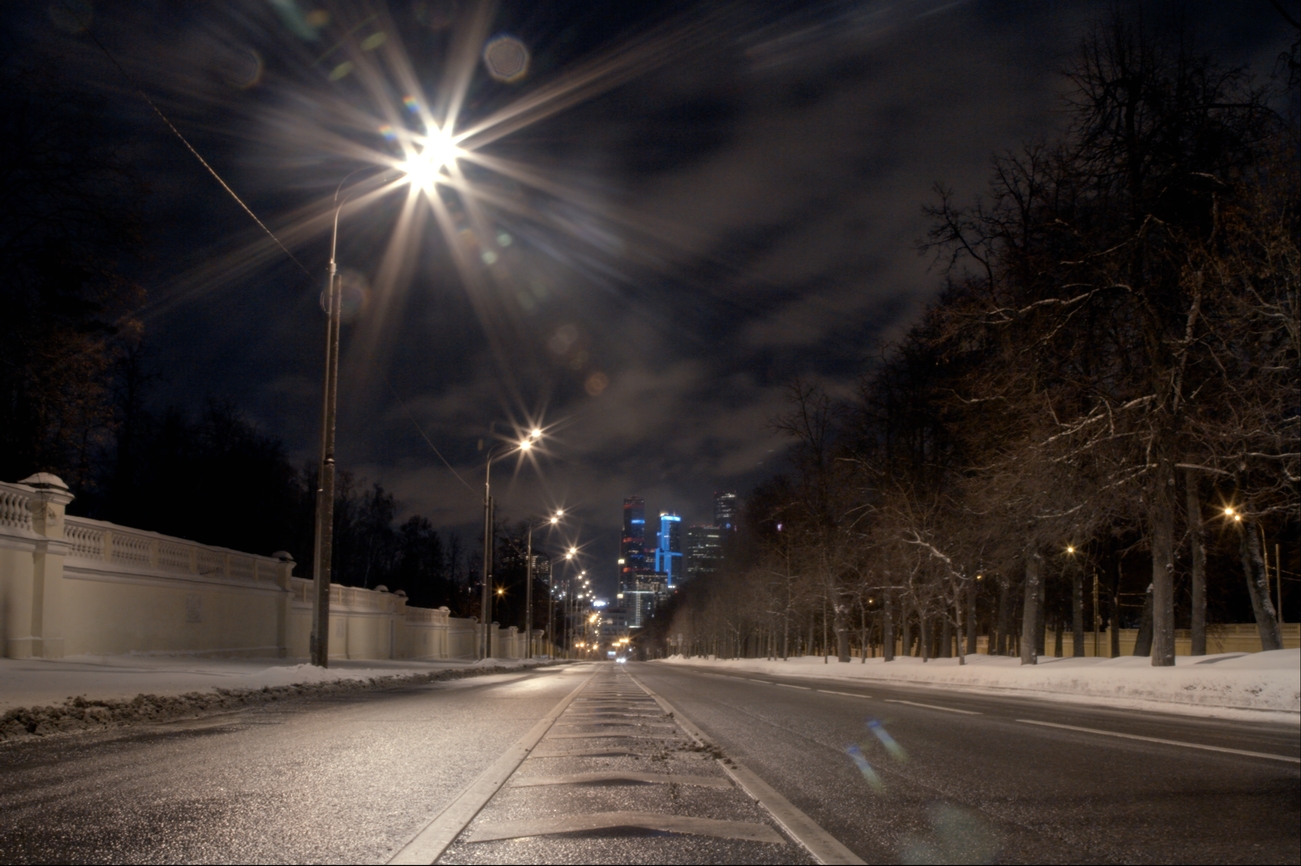}
        \includegraphics[width=\textwidth]{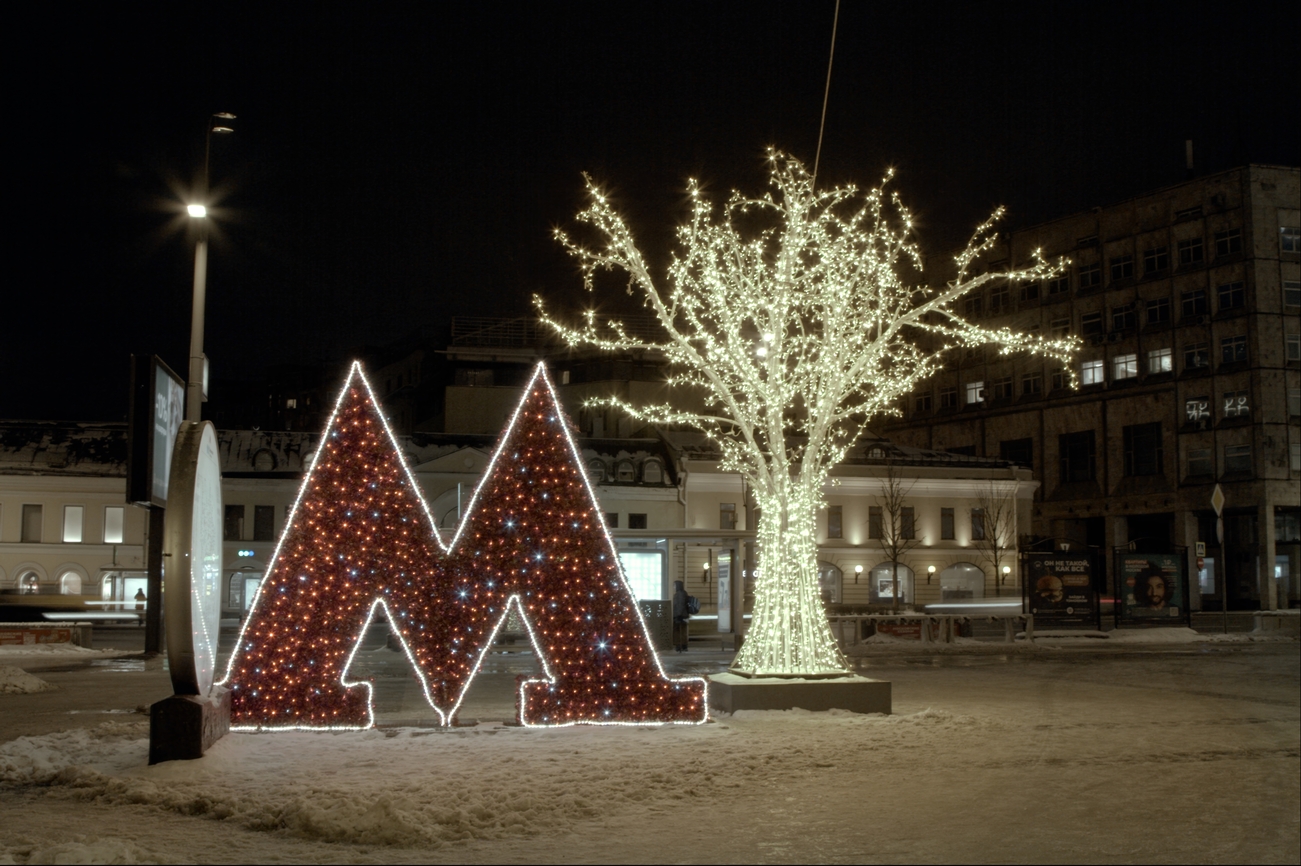}
        \includegraphics[width=\textwidth]{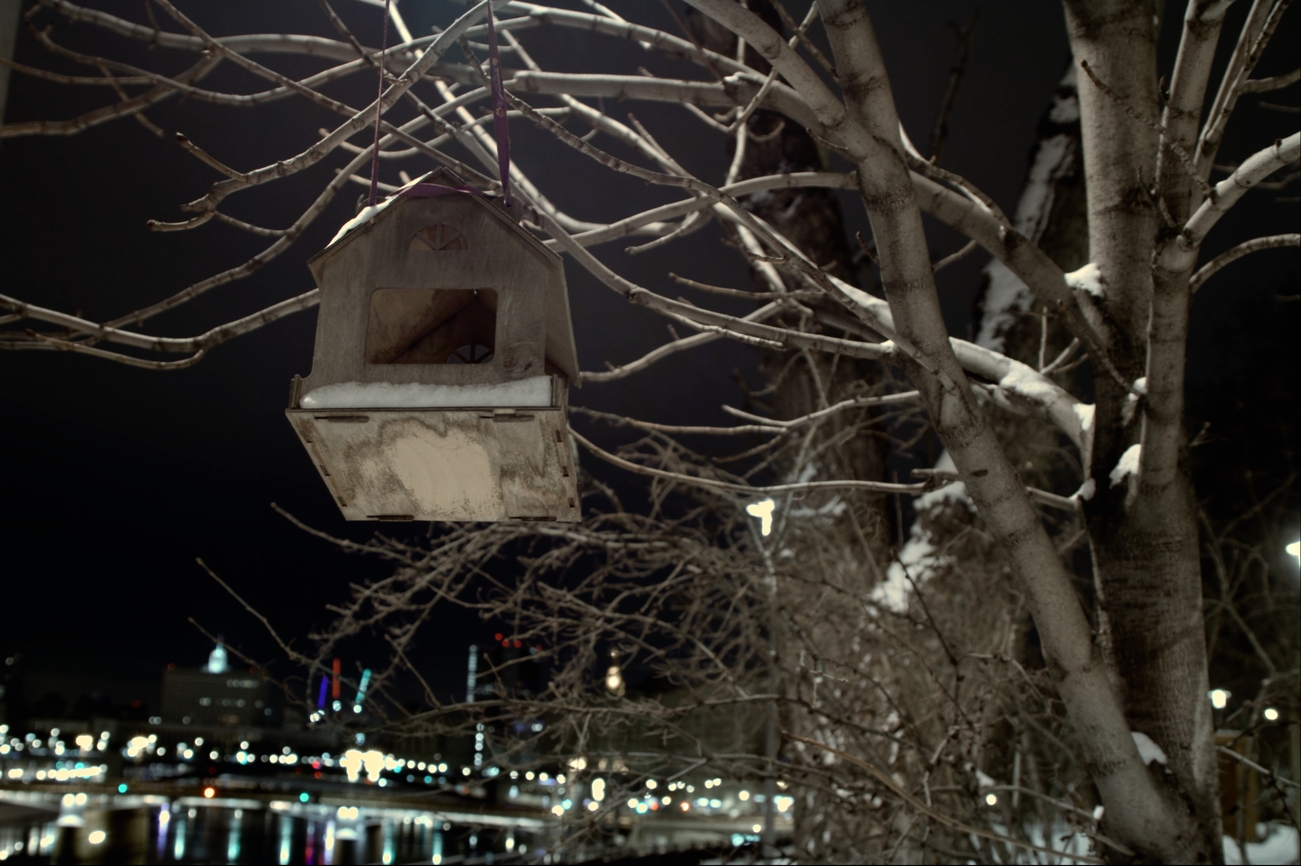}
        \caption{Style WB}
    \end{subfigure}
    \begin{subfigure}{0.24\textwidth}
        \includegraphics[width=\textwidth]{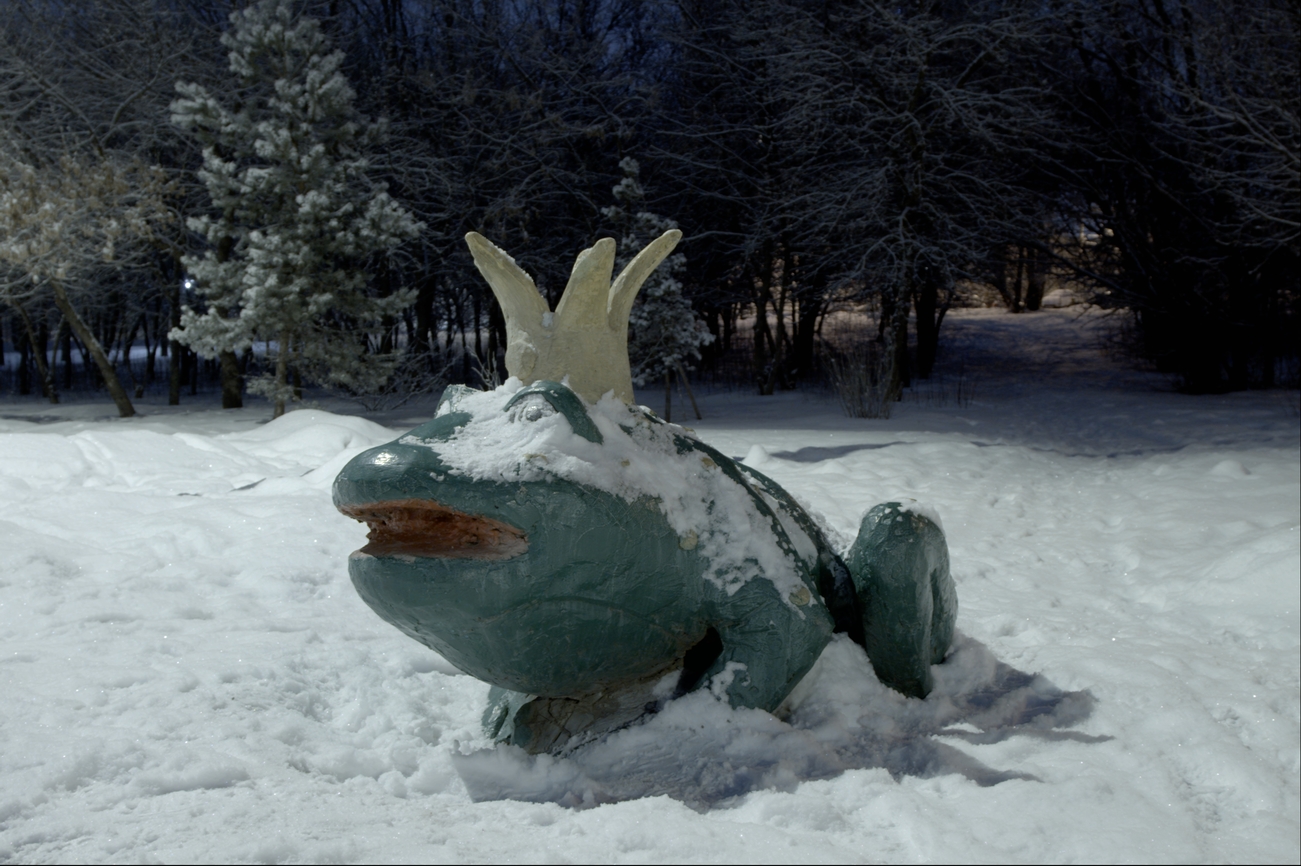} 
        \includegraphics[width=\textwidth]{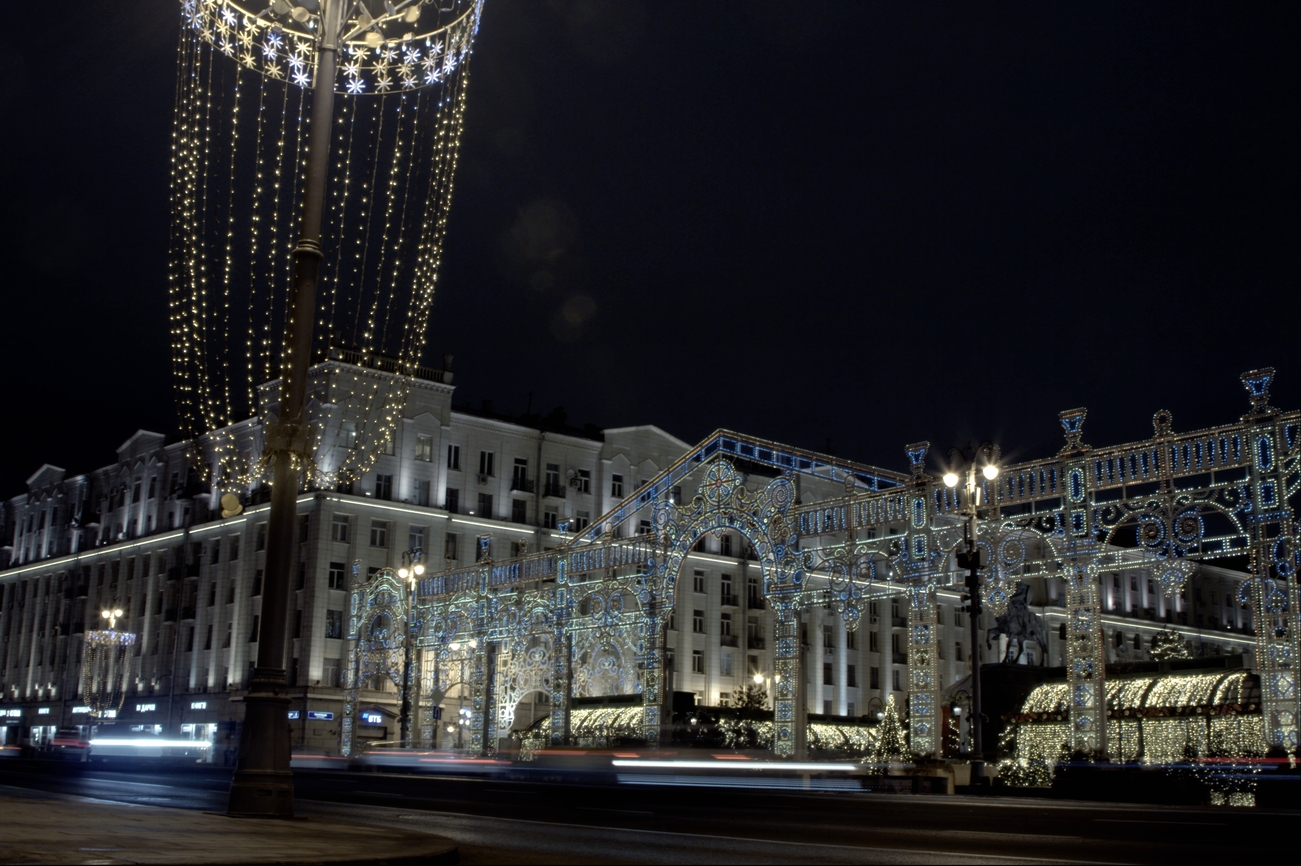}
        \includegraphics[width=\textwidth]{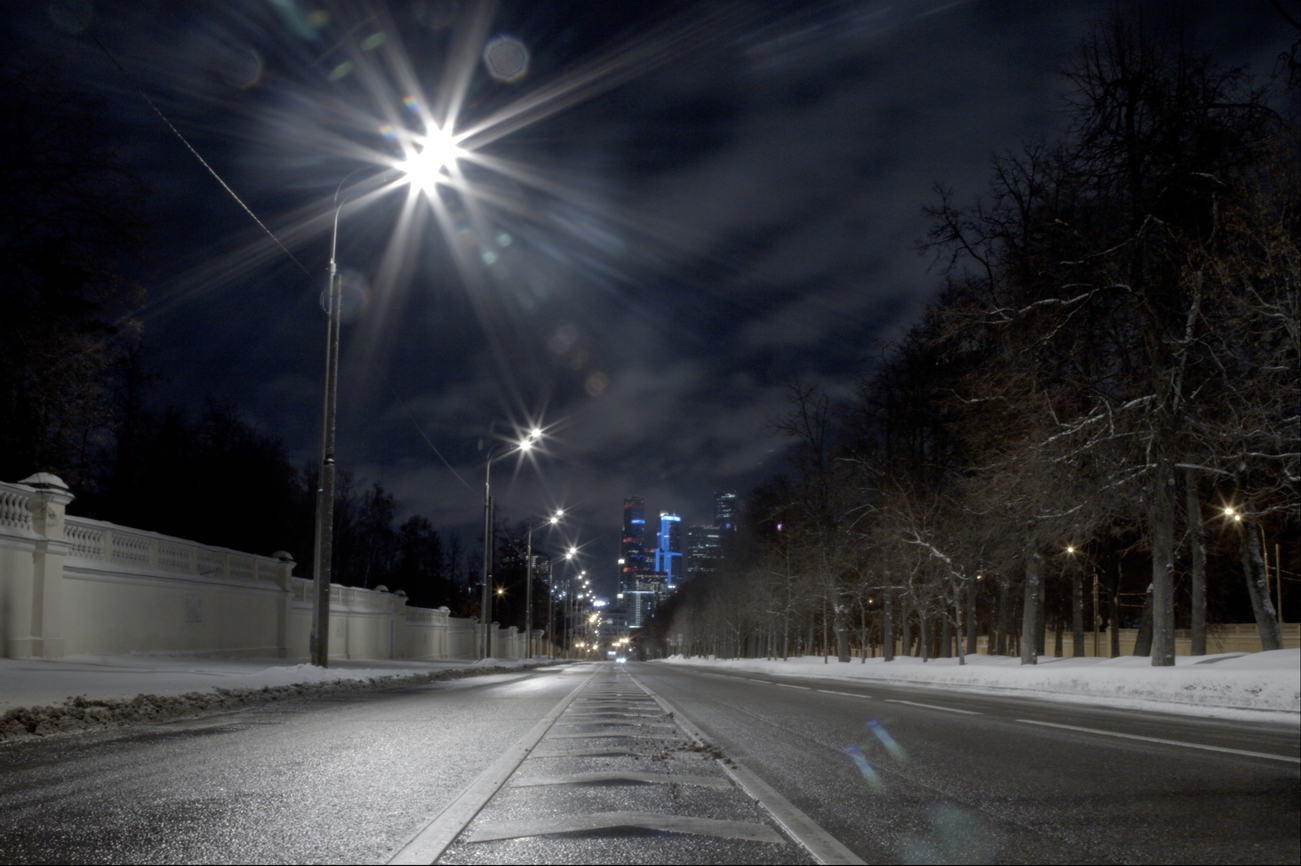} 
        \includegraphics[width=\textwidth]{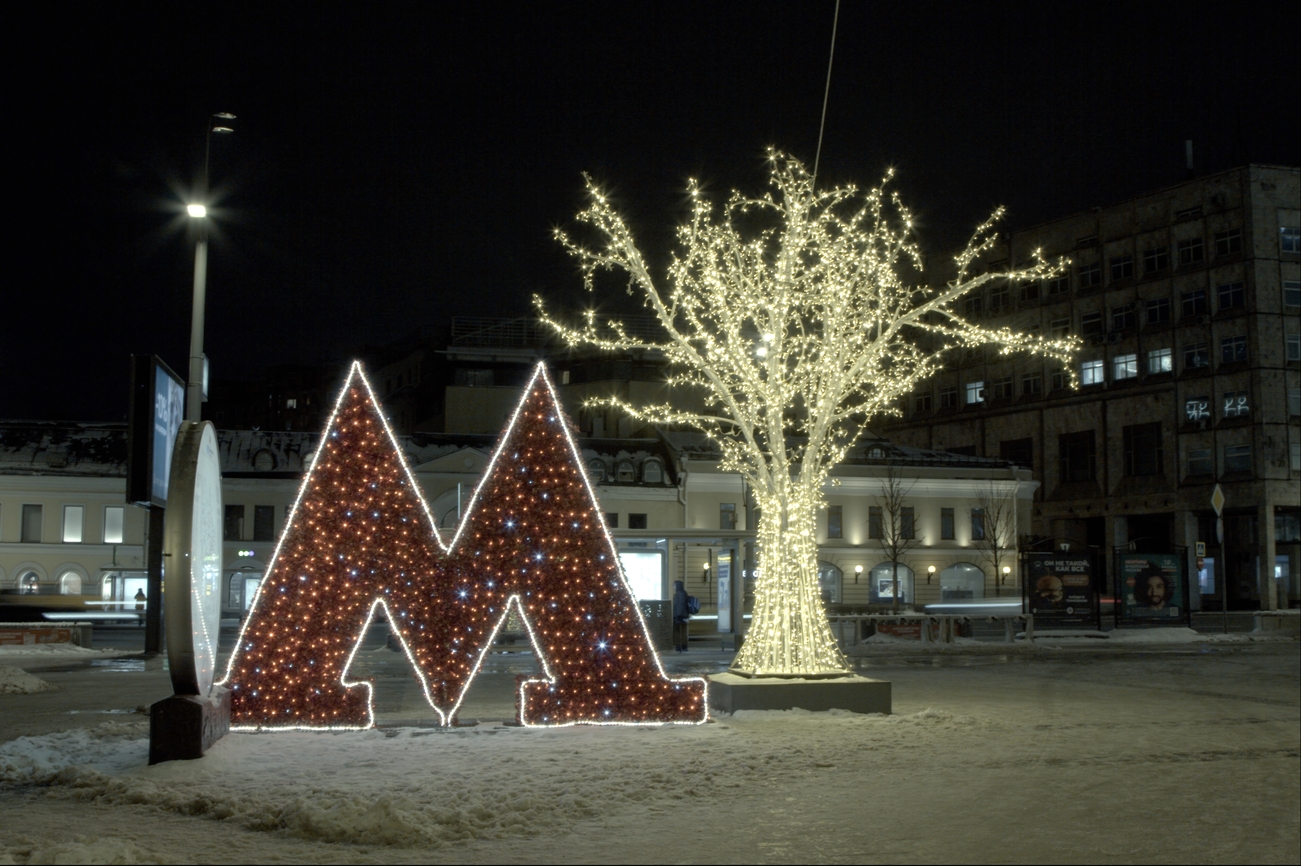}
        \includegraphics[width=\textwidth]{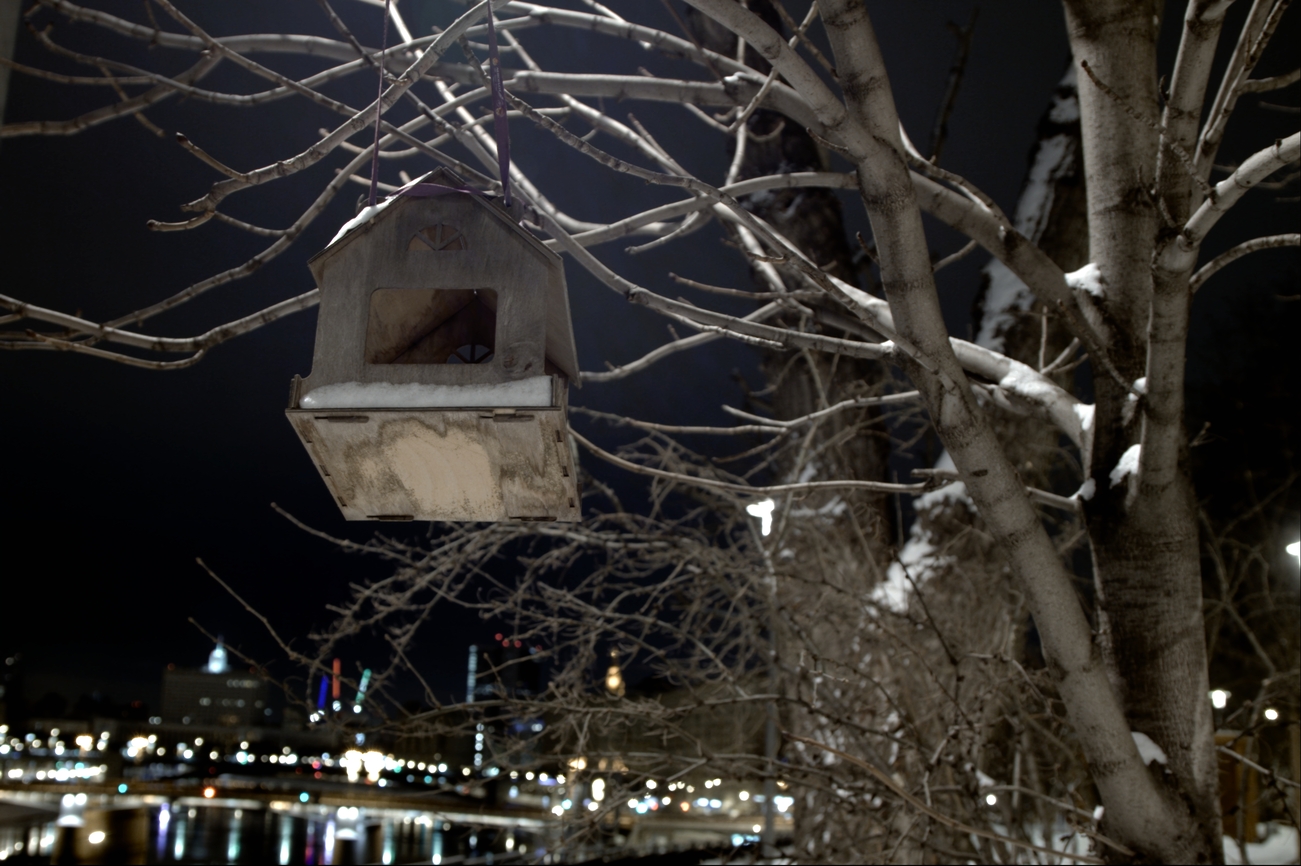} 
        \caption{DeNIM + Mixed WB}
    \end{subfigure}
    \begin{subfigure}{0.24\textwidth}
        \includegraphics[width=\textwidth]{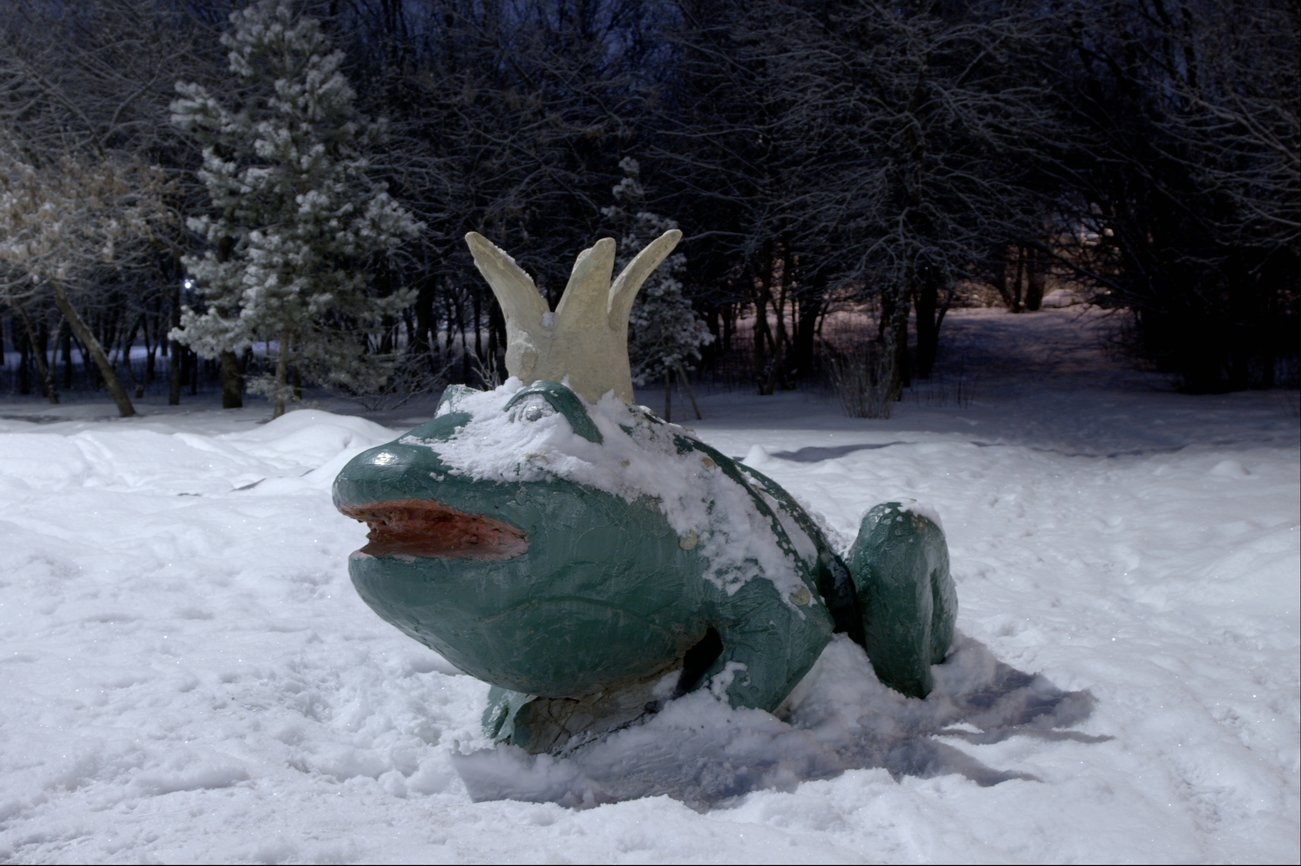} 
        \includegraphics[width=\textwidth]{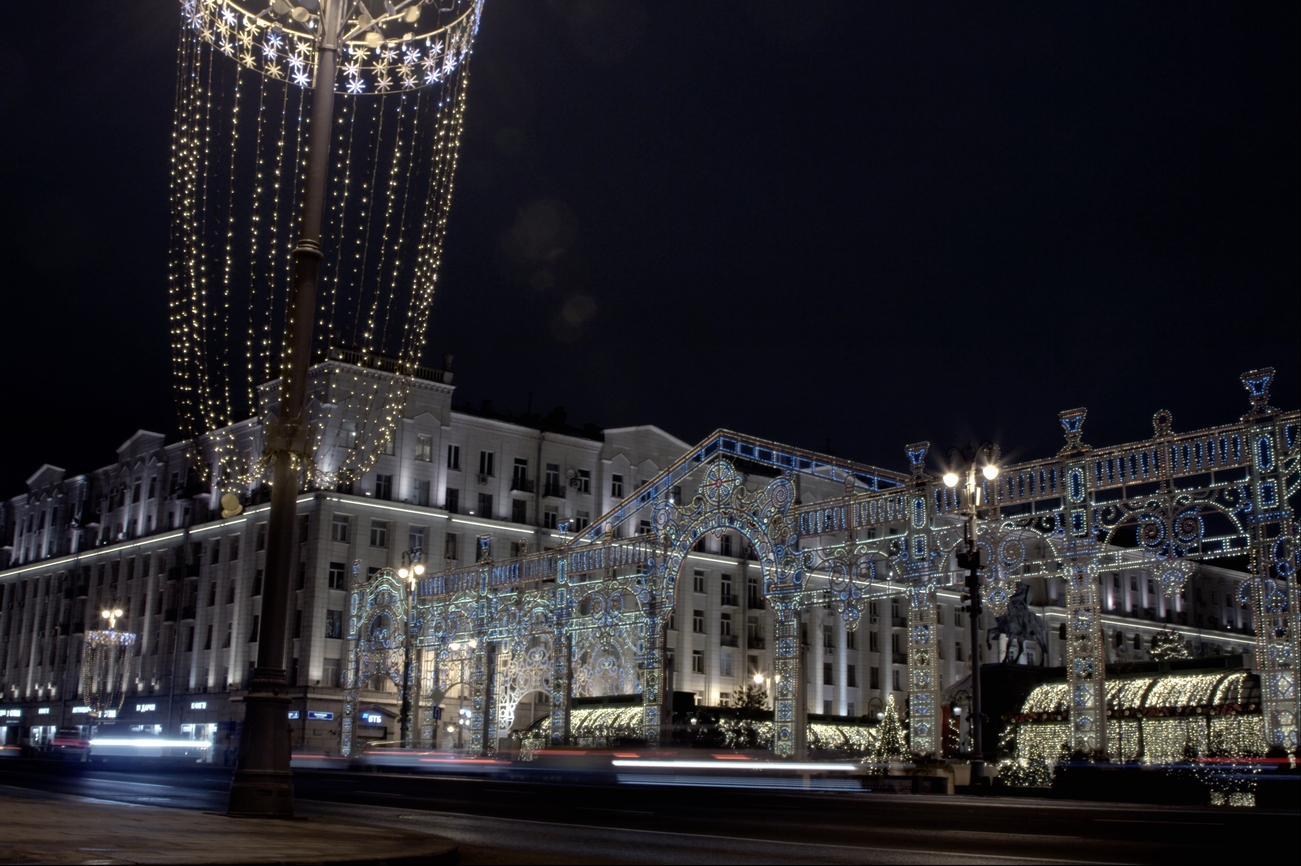}
        \includegraphics[width=\textwidth]{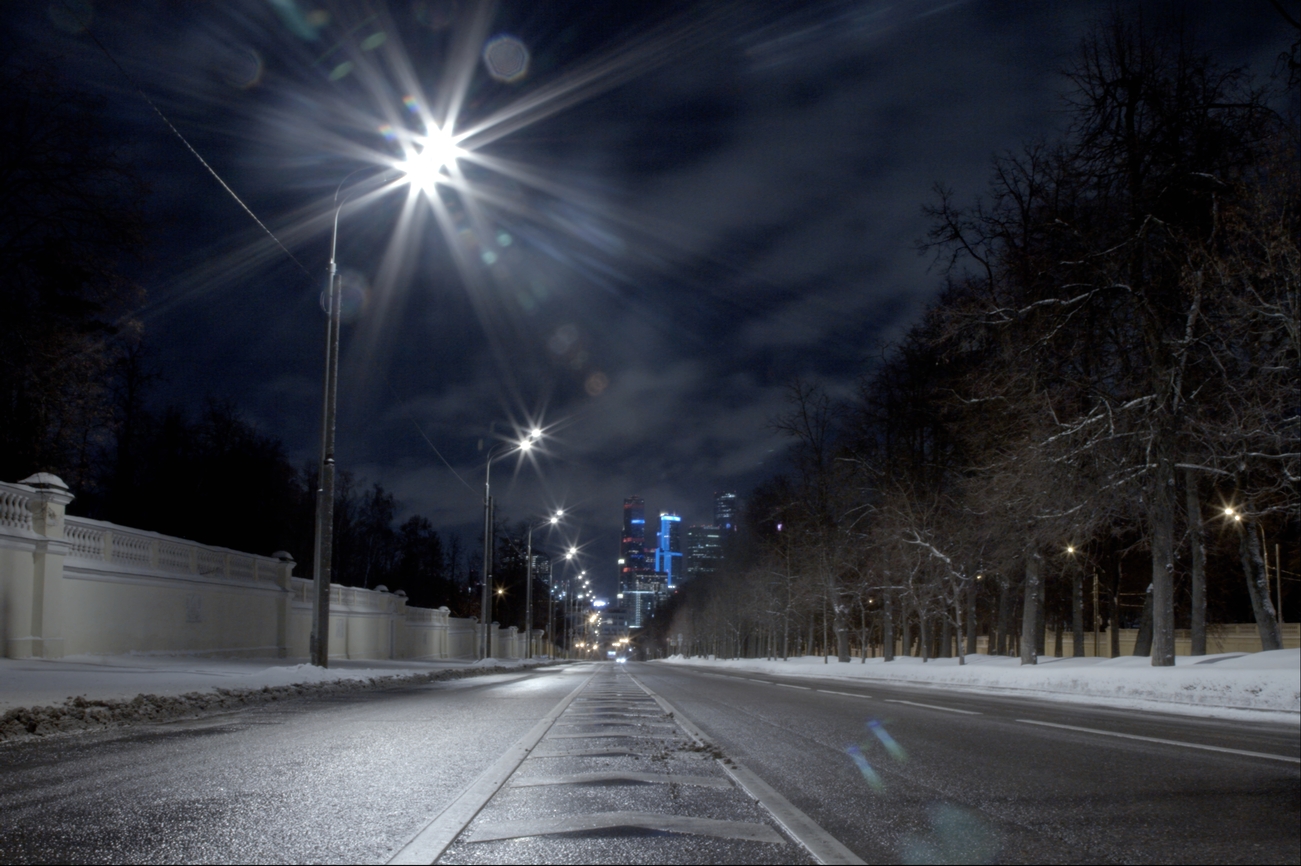} 
        \includegraphics[width=\textwidth]{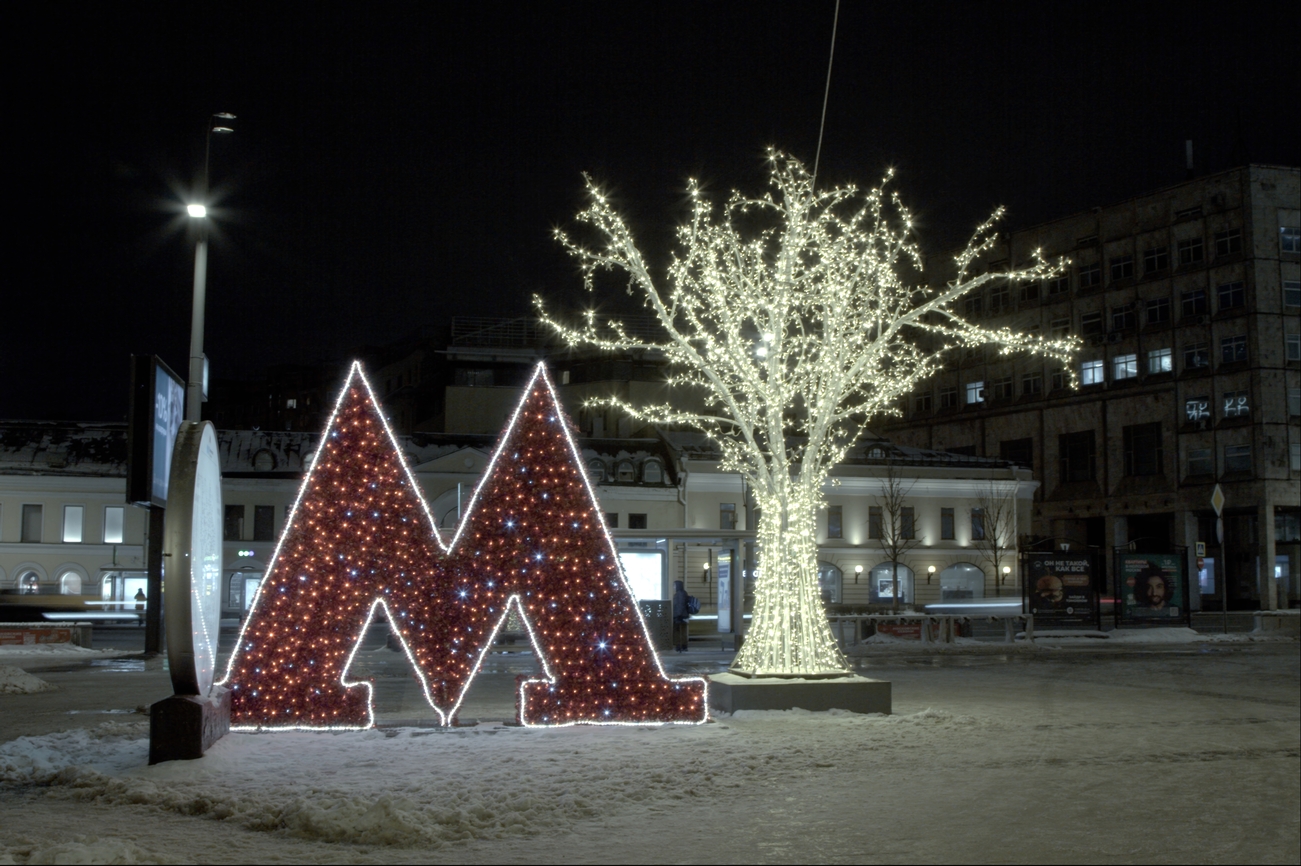}
        \includegraphics[width=\textwidth]{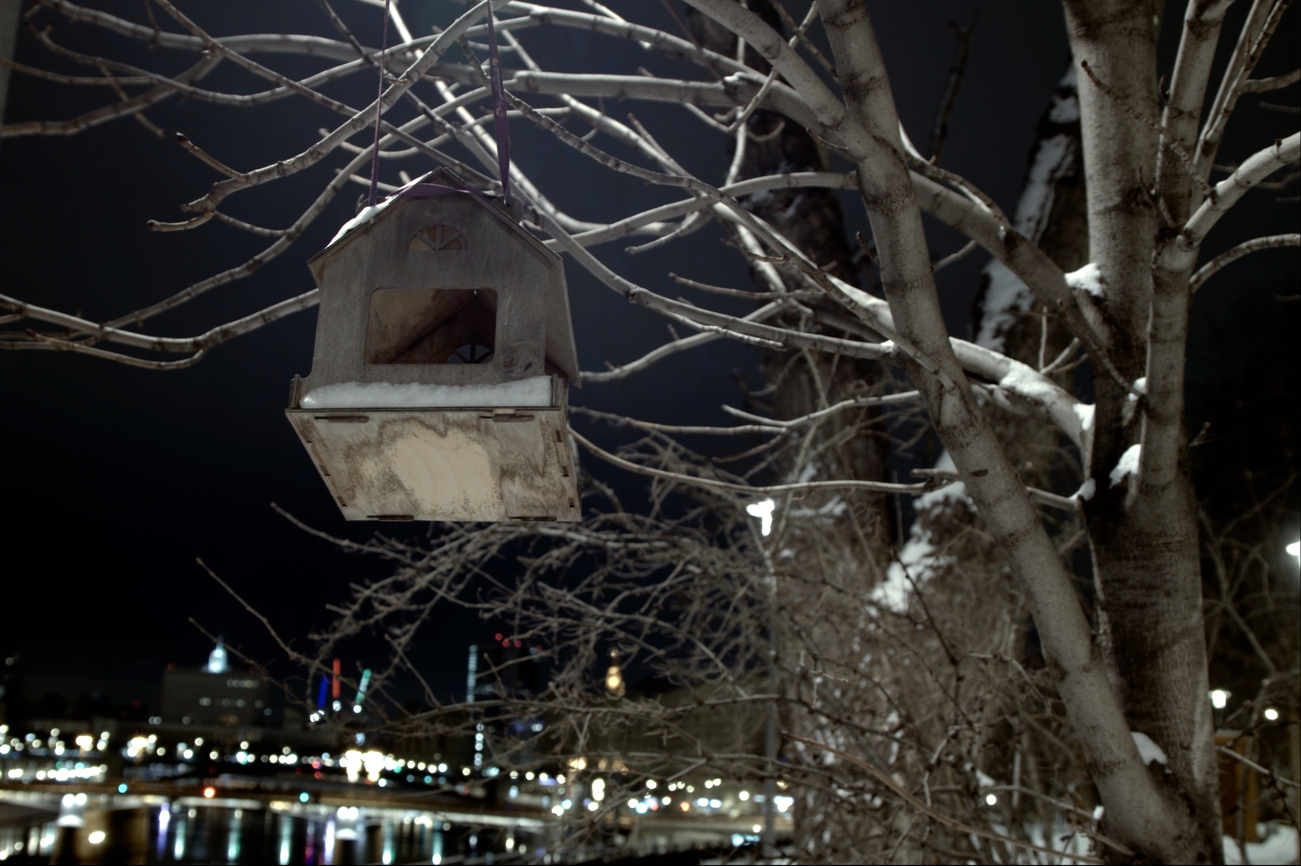}
        \caption{DeNIM + Style WB}
    \end{subfigure}
    \caption{Comparison of the night photography rendering results of our AWB correction strategy with Mixed WB \cite{Afifi_2022_WACV} and Style WB \cite{Kinli_2023_WACV} on the selected samples from Night Photography Rendering Challenge 23' evaluation set \cite{Shutova_2023_CVPR}. Image indices from top to bottom: $8678$, $8210$, $8817$, $8894$, $8941$.}\label{fig:qual-night} 
\end{figure*}

Night photography rendering \cite{Ershov_2022_CVPR, Shutova_2023_CVPR} is an emerging topic in digital imaging. Night image capturing poses significant challenges due to its inherent nature, characterized by low light conditions, diverse illuminant sources, and hardware limitations. In night image capturing, AWB correction plays a pivotal role in preserving the realistic perspective of the output, ensuring that it aligns with human perception and avoids distortions. As practiced in \cite{Kinli_2023_WACV}, we integrate our AWB correction strategy into the camera ISP for processing night images given in the evaluation part of Night Photography Challenge 23' \cite{Shutova_2023_CVPR}. In our pipeline, we incorporate the same operations, including gamma correction, tone mapping, auto-contrast, and denoising \cite{zamir2022restormer}, in the same order for all methods, but the only modification made is to the white-balancing strategies. Figure \ref{fig:qual-night} illustrates the rendering results of various camera pipeline variants that encompass the prior works and our proposed strategy as the AWB correction method. The rendering results demonstrate that our strategy effectively produces more natural night images by mitigating undesired color casts commonly encountered in real-world scenarios.

\noindent\textbf{Quantitative evaluation:}
The benchmark on single-illuminant Cube+ dataset \cite{global_tonal_adj} is presented in Table \ref{table:results_1}. Following the same experimental setup in the prior works \cite{Afifi_2022_WACV, Kinli_2023_WACV}, we have used two different patch sizes for the backbone network (\textit{i.e.}, $64$ and $128$), and we designed the input image with two sets of WB settings where the default choices include Tungsten, Daylight, and Shade, while we further incorporate Fluorescent and Cloudy color temperatures to enhance the versatility of the method. The quantitative results indicate that our strategy achieves not only increasing efficiency but also improving performance across all different patch sizes and WB settings, as evidenced by all evaluation metrics. The main observations extracted from these results are as follows: (1) In contrast to the results obtained with Style WB, the best-performing variant appears to be when using a patch size of 64 and incorporating all possible WB settings. This configuration leads to superior performance when compared to other settings. (2) The notable increase in performance, specifically observed on the third quantiles of all evaluation metrics, deserves highlighting. This observation suggests that our strategy can produce more robust results, particularly when dealing with challenging samples.
\begin{figure*}[h!]
    \centering
    \begin{subfigure}{0.24\textwidth}
        \includegraphics[width=\textwidth]{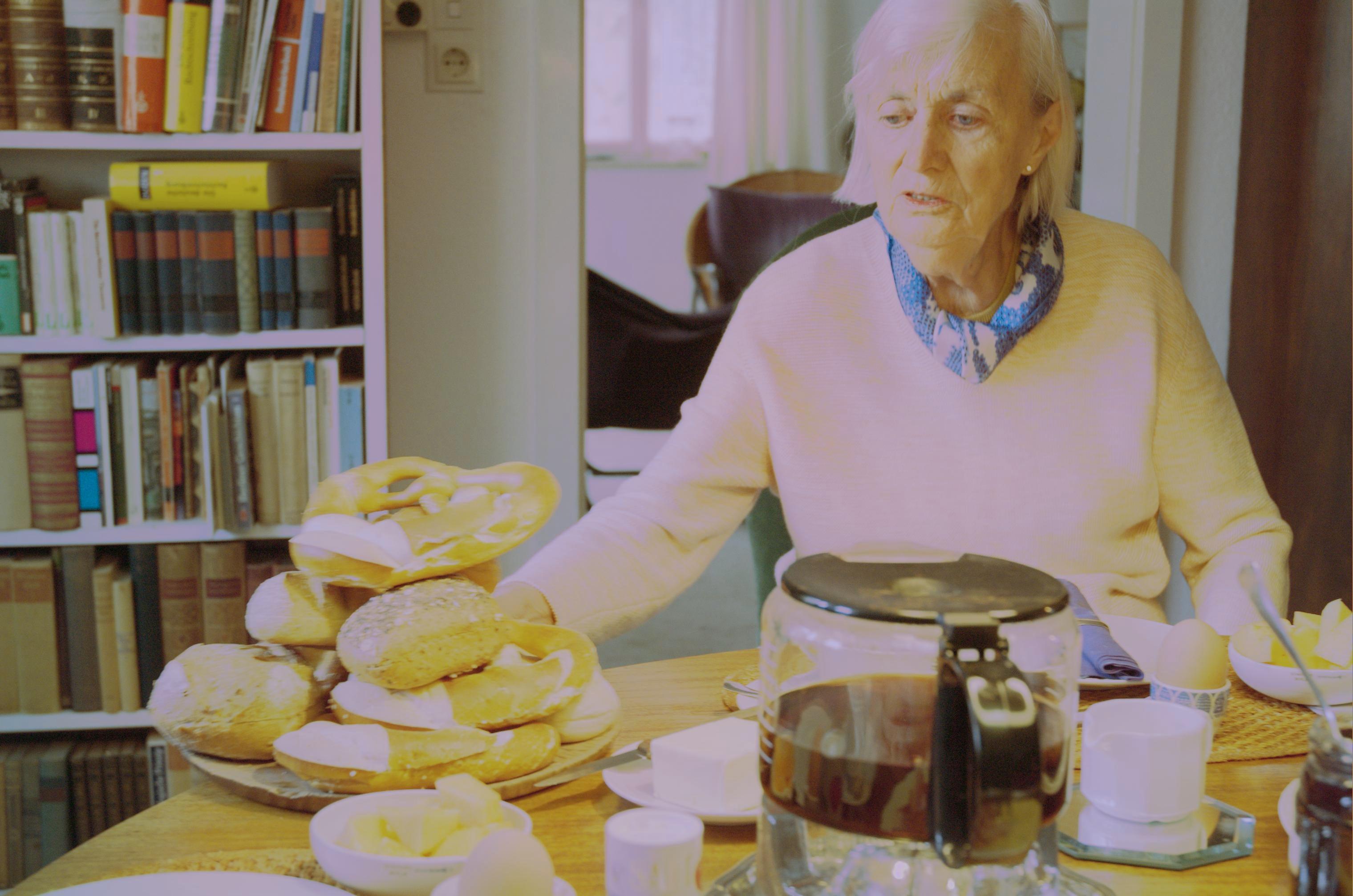} 
        \caption{Mixed WB}
    \end{subfigure}
    \begin{subfigure}{0.24\textwidth}
        \includegraphics[width=\textwidth]{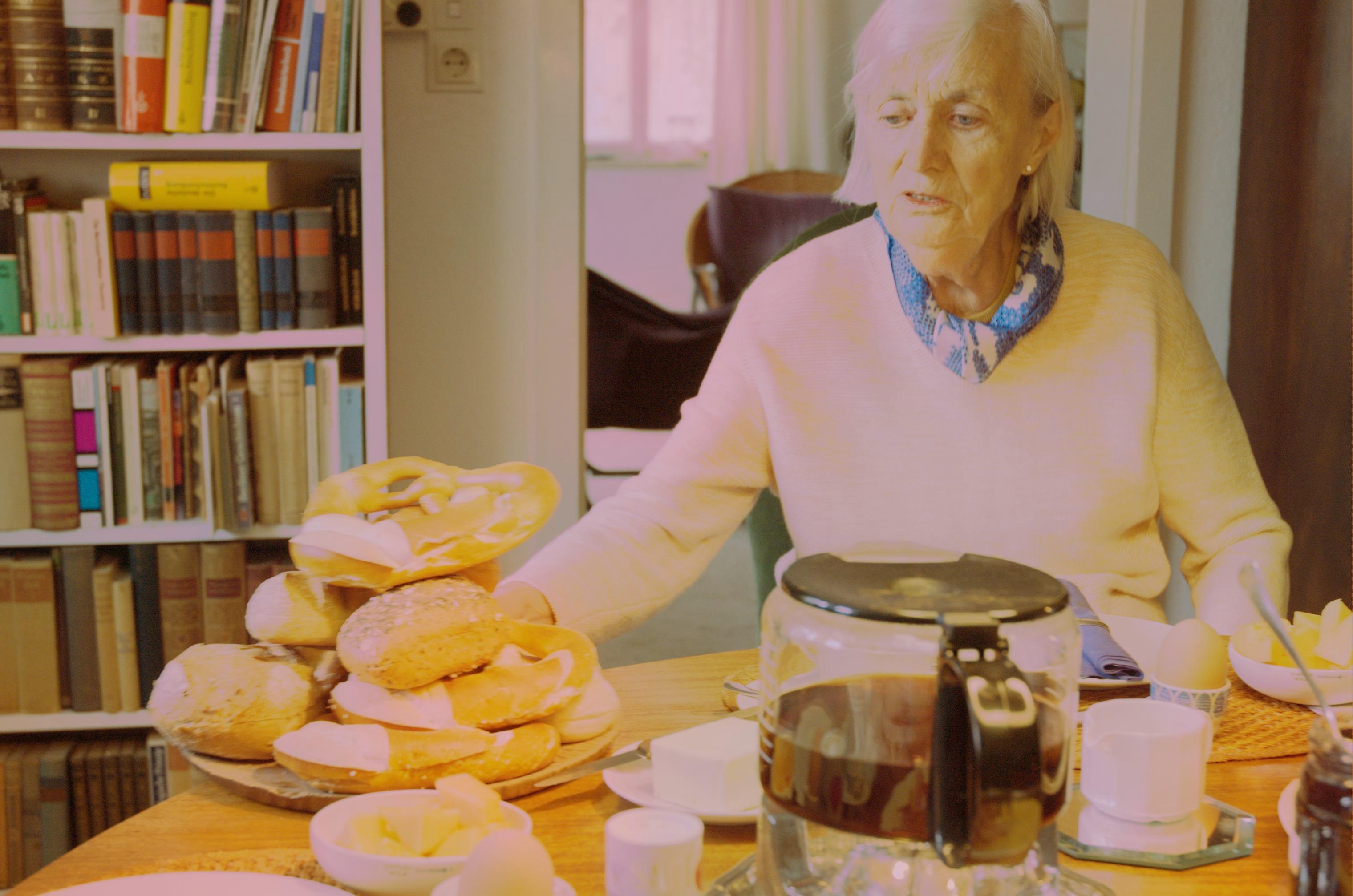} 
        \caption{Style WB}
    \end{subfigure}
    \begin{subfigure}{0.24\textwidth}
        \includegraphics[width=\textwidth]{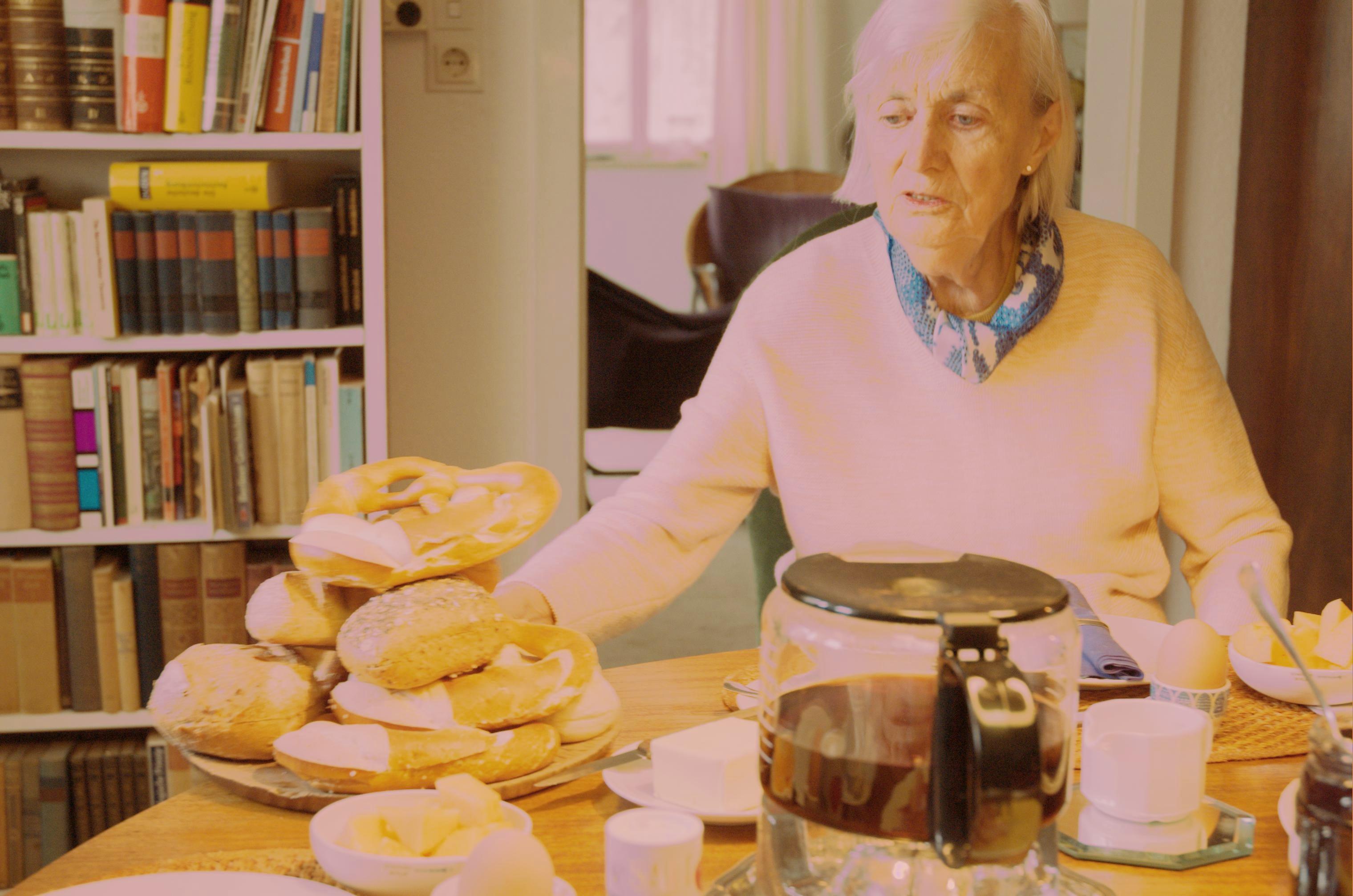} 
        \caption{DeNIM + Mixed WB}
    \end{subfigure}
    \begin{subfigure}{0.24\textwidth}
        \includegraphics[width=\textwidth]{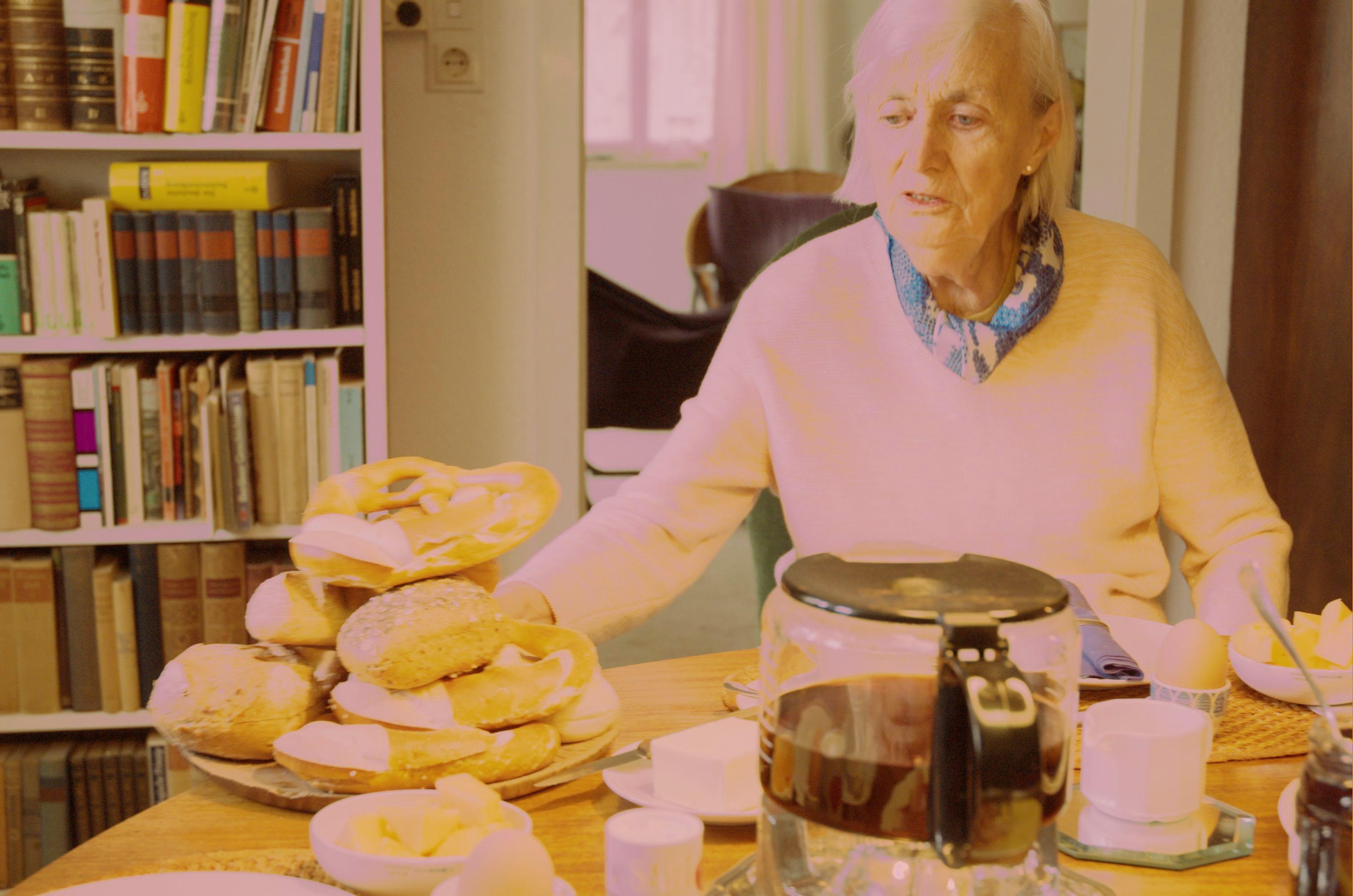} 
        \caption{DeNIM + Style WB}
    \end{subfigure}
    \caption{Failing to address unrealistic color casts, and not effectively handled by any AWB correction methods on MIT-Adobe FiveK dataset \cite{global_tonal_adj}.}
    \label{fig:limitations}
\end{figure*}
(3) Confirming the findings in \cite{Afifi_2022_WACV,Kinli_2023_WACV}, we observe that smaller patch sizes tend to lead to better modeling of the illuminant, and in our case, also learning color mappings. (4) We encountered difficulties in identifying a consistent pattern for the mean-squared error (MSE) metric when compared to the other two metrics, and this may suggest that MSE may not adequately capture the quality of color correction achieved by the different methods. We believe that this particular metric might not be suitable for accurately measuring the performance of AWB correction.

\noindent\textbf{Efficiency:}
The results presented in Table \ref{table:results_2} demonstrate the efficiency of our proposed strategy when compared to its prior works across different criteria. Specifically, we evaluated the efficiency based on the following criteria: the processing time (Time (s)), the model complexity in terms of parameter count (Param (M)), and computational load measured in Floating Point Operations Per Second (FLOPS (G)). In terms of processing time, our strategy significantly reduces the time required to process the images for AWB correction. The reduction in processing time is accomplished by designing a model that allows discarding the post-processing operations (\textit{i.e.}, multi-scale inference, and edge-aware smoothing) and adopting simple learnable projection matrices in place of the decoder. DeNIM shows a remarkable speed advantage, being at least 35 times faster than previous models (up to 1700 times faster when post-processing is included). 

Next, the model complexity is an essential factor to consider. DeNIM leads to a slight increase in the number of parameters compared to the prior works, even though it discards the decoder of the baseline models. The reason behind the increasing number of parameters lies in the decision to use fully-connected layers as projection matrices, as opposed to convolutional layers in the decoder. Fully-connected layers require more parameters, due to their dense connections between all input and output neurons. This design choice may have led to a slightly higher model complexity, however, it is important to note that this decision does not significantly impact the processing time. Lastly, we measure the computational load of all methods in terms of FLOPS. Lower FLOPS values imply less computational resources required, hence better efficiency. When DeNIM is trained with the Mixed WB backbone, it achieves a remarkable reduction in FLOPS by approximately 97\%. Similarly, when trained with the Style WB backbone, the FLOPS are reduced by approximately 65\%. This substantial decrease in computational load highlights the remarkable efficiency of our strategy compared to the prior works.

\noindent\textbf{Limitations:}
Although deep-learning-driven AWB methods generally demonstrate significant resilience across various different scenarios, there are occasional examples where they yield unsatisfactory results. As shown in Figure \ref{fig:limitations}, AWB correction operations may fail to address unrealistic color casts and produce poor results which do not align with human visual perception. At this point, our strategy also may not be able to handle the challenges effectively, primarily since it relies on the feature extraction part of the prior models. We can state that it may struggle to address certain complex and uncommon scenarios, which leads to sub-optimal results. Moreover, to further investigate the performance in handling more challenging cases, our strategy can be tested on multi-illuminant datasets \cite{Afifi_2022_WACV, kim2021large}. By subjecting this strategy to such datasets, we can gain valuable insights into its capabilities and limitations in handling diverse and complex lighting scenarios, and we left it as future work.

\section{Conclusion}

In this paper, we have introduced a novel and efficient deep learning-based AWB correction strategy built on top of the current state-of-the-art methods. This strategy incorporates the idea of deterministic color mapping by leveraging the encoder of existing AWB models and learnable projection matrices. Through extensive experiments, we showed the effectiveness of our strategy by achieving at least 35 times faster processing while surpassing the performance of state-of-the-art methods on high-resolution images. Our research provides a promising solution for real-time, high-quality color correction in practical scenarios, even in digital camera chipsets, addressing the challenges posed by increasing model complexities for better performance.

{\small
\bibliographystyle{ieee_fullname}
\bibliography{egpaper_for_review}
}

\end{document}